%% file: main.tex
\PassOptionsToPackage{table}{xcolor}
\PassOptionsToPackage{noend}{algorithmic}
\documentclass{article}

\usepackage{microtype}
\usepackage{graphicx}
\usepackage{subfigure}
\usepackage{booktabs} 

\usepackage{hyperref}

\usepackage[accepted]{icml2024}

\usepackage{amsmath}
\usepackage{amssymb}
\usepackage{mathtools}
\usepackage{amsthm}

\usepackage[capitalize,noabbrev]{cleveref}

\theoremstyle{plain}
\newtheorem{theorem}{Theorem}[section]

\newtheorem{lemma}[theorem]{Lemma}

\theoremstyle{definition}
\newtheorem{definition}[theorem]{Definition}
\newtheorem{assumption}[theorem]{Assumption}
\theoremstyle{remark}
\newtheorem{remark}[theorem]{Remark}

\usepackage[textsize=tiny]{todonotes}

\icmltitlerunning{Transformers on Markov data}

\usepackage{ulem}
\usepackage{caption}
\usepackage{subcaption}
\usepackage{stackengine}
\newcommand\xrowht[2][0]{\addstackgap[.5\dimexpr#2\relax]{\vphantom{#1}}}

\usepackage{tikz}
\usetikzlibrary{calc}
\usetikzlibrary{shapes.geometric}
\usetikzlibrary{arrows.meta}
\usetikzlibrary{positioning}
\usetikzlibrary{decorations.markings}
\usepackage[utf8]{inputenc} 
\usepackage[T1]{fontenc}    
\usepackage{hyperref}       
\usepackage{url}            
\usepackage{booktabs}       
\usepackage{amsfonts}       
\usepackage{nicefrac}       
\usepackage{microtype}      
\usepackage{xcolor}         
\usepackage{bm}

\usepackage{pgfplots}
\usepackage{colortbl}

\usepackage{newfloat}
\DeclareFloatingEnvironment[
    fileext=los,
    listname={Architectures},
    name=Architecture,
    placement=tbhp,
    within=none,
]{Architecture}
\input{commands_icml}
\usepackage[capitalize,noabbrev]{cleveref}
\crefname{assumption}{assumption}{assumptions}

\begin{document}

\twocolumn[
\icmltitle{Transformers on Markov Data: Constant Depth Suffices}



\icmlsetsymbol{equal}{*}

\begin{icmlauthorlist}
\icmlauthor{Nived Rajaraman}{yyy}
\icmlauthor{Marco Bondaschi}{xxx}
\icmlauthor{Kannan Ramchandran}{yyy}
\icmlauthor{Michael Gastpar}{xxx}
\icmlauthor{Ashok Vardhan Makkuva}{xxx}
\end{icmlauthorlist}

\icmlaffiliation{yyy}{UC Berkeley}
\icmlaffiliation{xxx}{EPFL}

\icmlcorrespondingauthor{Nived Rajaraman}{nived.rajaraman@berkeley.edu}

\icmlkeywords{Transformers, Markov chains, Interpretability}

\vskip 0.3in
]



\printAffiliationsAndNotice{} 

\begin{abstract}
Attention-based transformers have been remarkably successful at modeling generative processes across various domains and modalities. In this paper, we study the behavior of transformers on data drawn from \kth Markov processes, where the conditional distribution of the next symbol in a sequence depends on the previous $k$ symbols observed. We observe a surprising phenomenon empirically which contradicts previous findings: when trained for sufficiently long, a transformer with a fixed depth and $1$ head per layer is able to achieve low test loss on sequences drawn from \kth Markov sources, even as $k$ grows. Furthermore, this low test loss is achieved by the transformer's ability to represent and learn the in-context conditional empirical distribution. On the theoretical side, our main result is that a transformer with a single head and three layers can represent the in-context conditional empirical distribution for \kth Markov sources, concurring with our empirical observations. Along the way, we prove that \textit{attention-only} transformers with $O(\log_2(k))$ layers can represent the in-context conditional empirical distribution by composing induction heads to track the previous $k$ symbols in the sequence. These results provide more insight into our current understanding of the mechanisms by which transformers learn to capture context, by understanding their behavior on Markov sources. Code is available at: \url{https://github.com/Bond1995/Constant-depth-Transformers}
\end{abstract}













\doparttoc
\faketableofcontents

\section{Introduction}

Attention-based transformers have revolutionized the field of natural language processing (NLP) \cite{vaswani2017attention,brown2020language} and beyond \cite{dosovitskiy2021image,he2021masked}, achieving significant performance gains across tasks like machine translation, text generation, and sentiment analysis. A key factor in their success is their ability to model sequences far more efficiently, and the ability to learn in-context \cite{bietti2023birth,edelman2024evolution}.

To understand this capability, a canonical approach is to sample the input from a {\it \kth Markov process}, where the next symbol's conditional distribution depends only on the previous $k$ symbols. Recent studies \cite{makkuva2024attention,edelman2024evolution,nichani2024transformers} have investigated the ability of transformers to learn Markov processes and establish that learning happens in phases. The transformer eventually learns to represent the conditional $k$-gram model, which is the in-context MLE of the Markov process.

The results in \cite{edelman2024evolution,nichani2024transformers} seem to suggest that for low depth transformers to learn Markov processes of order $k$, it is essential that the number of heads scale linearly in $k$. At first glance, this is a bit concerning - real world data generating processes often contain long-range dependencies. How is it that transformers succeed at capturing these kinds of long-range dependencies, while at the same time requiring so many heads to be able to capture the necessary context for \kth Markov sources.

To understand the nature of this phenomenon, we train low-depth transformers on \kth Markov sources. These experiments result in two surprising empirical phenomena that seem to contradict previous findings: when trained for sufficiently long, $(i)$ a $2$-layer, $1$-head transformer can learn \kth Markov processes for $k$ as large as $4$, $(ii)$ a $3$-layer, $1$-head transformer is able to achieve low test loss on sequences drawn from \kth Markov sources, even as $k$ grows to be as large as $8$ (\prettyref{fig:order8}). In both cases, the values of $k$ for which the models appear to learn \kth Markov sources are much higher than those predicted in prior experiments \cite{edelman2024evolution,nichani2024transformers}. This discrepancy shows that our understanding of the mechanisms used by transformers to learn \kth Markov processes is not complete and raises a broader question: 
\begin{quote}
    \it{What is the interplay between depth, number of heads and non-linearity in learning \kth Markov processes?}
\end{quote}
In this paper, we approach this question from the point of view of representation power, and provide some partial explanations toward the phenomena illustrated previously.  



\begin{figure}
    \centering
    \begin{tikzpicture}
    \node at (-1.75, 0) {$\cdots$};
    \foreach \i/\j in {-2/2, -1.5/2, -1/3, -0.5/1, 0/2, 0.5/2, 1/3, 1.5/0, 4.5/{\footnotesize $X_{n+1}$}} {
        \node[fill=gray!25, minimum size=0.5cm] (box\i) at (1+1*\i, 0) {\j};
    }
    \foreach \i/\j in {2/1, 2.5/0, 3/1, 3.5/0} {
        \node[minimum size=0.5cm, fill=red!25] (boxb\i) at (1+1*\i, 0) {\j};
    }
\end{tikzpicture}
    \caption{\kth Markov processes for $k=4$. The next symbol $X_{n+1}$ in the sequence is sampled from the distribution $P(\cdot | X_n, X_{n-1}, X_{n-2}, X_{n-3})$ which only depends on the last $k (=4)$ symbols (marked in red).}
    \label{fig:markov}
\end{figure}

Our main contributions are as follows:

\begin{enumerate}
    \item We show, rather surprisingly, that the standard transformer architecture with $3$ layers and $1$ head per layer is capable of representing the conditional $k$-gram model (\Cref{def:condK}), and thereby learn \kth Markov models in-context.
    \item Along the way to building up to this result,  we consider the simpler family of \textit{attention-only transformers} and show that they can represent the conditional $k$-gram model with $\lceil \log_2 (k+1) \rceil$ layers.
    \item Under a natural assumption on the nature of the attention patterns learnt by the transformer, we then argue that for $k \ge 3$ attention-only transformers \textit{need} at least $\lceil 1+\log_2 (k-2) \rceil$ layers to represent a ``\kth induction head'' (\Cref{def:k-order_IH}). Empirically, transformers are observed to learn \kth induction heads whenever they achieve small test error \cite{edelman2024evolution}.
\end{enumerate}

The last result is a consequence of a more general tradeoff between the number of layers, $L$, and heads per layer, $H$, an attention-only transformer requires to represent a \kth induction head, under a natural assumption on the learnt attention patterns. In conjunction, these results also reveal the role of non-linearities (aside from the softmax in the attention) in the transformer architecture. In particular, it appears that layer normalization plays a critical role in the ability of constant-depth transformers to learn the conditional $k$-gram model. Together with the experimental results mentioned previously, these results paint a more comprehensive picture about the representation landscape of transformers in the context of \kth Markov processes.

\begin{table}[]
    \centering
    \begin{tabular}{|c|p{1.9em}|c|c|}
    \rowcolor{gray!50}
        \hline
        \cellcolor{blue!25} \xrowht[()]{6pt} & \multicolumn{2}{c|}{Attention-only} & Standard \\
        \hline
        Layers $(L)$ \xrowht[()]{6pt} & \centering $2$ & $\lceil \log_2(k+1) \rceil$ & $3$ \\
        \hline
        Heads $(H)$ \xrowht[()]{6pt} & \centering $k$ & $1$ & $1$ \\
        \hline
    \end{tabular}
    \vspace{1em}
    \caption[Caption for LOF]{Each column in this table indicates that there is a transformer with $L$ layers and $H$ heads in the first layer which can represent the conditional $k$-gram model.\footnotemark}
    \label{tab:results}
\end{table}


\paragraph{Notation.} 
Scalars are denoted by italic lower case letters like $x,y$ and Euclidean vectors and matrices in bold $\bx, \by, \bM$, etc. 
The notation $\bm{0}_{p \times q}$ (resp. $\bm{1}_{p \times q}$) refers to the all-zero (resp. all-one) matrix. When it is clear from the context, we omit the dimensions of a matrix. 
Define $[S] \define \{1,2,\ldots, S \}$ for $S \in \naturals$. $\mathbb{I} (\cdot)$ denotes the indicator function and $\mathrm{Unif} (S)$ denotes the uniform distribution over a set $S$.

\footnotetext{The requisite embedding dimension and bit-precision to achieve a target additive approximation is discussed in more detail in \Cref{sec:logk,sec:constant}}
\subsection{Related work}

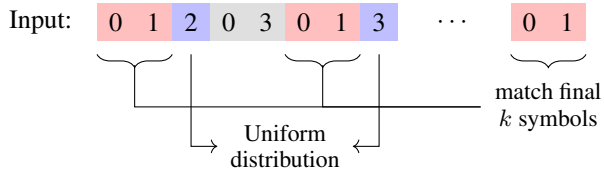
\begin{figure}
    \centering
    \begin{tikzpicture}
    \node[align=left] (ins) at (-1,0) {Input:};
    \foreach \i/\j in {0/0, 0.5/1, 2.5/0, 3/1} {
        \node[minimum size=0.5cm, fill=red!25] (box\i) at (\i, 0) {\j};
    }
    \foreach \i/\j in {1.5/0, 2/3} {
        \node[minimum size=0.5cm, fill=gray!25] (box\i) at (\i, 0) {\j};
    }
    \foreach \i/\j in {1/2, 3.5/3} {
        \node[minimum size=0.5cm, fill=blue!25] (box\i) at (\i, 0) {\j};
    }
    
    \node at (4.5, 0) {$\cdots$};

    \foreach \i/\j in {0/0, 0.5/1} {
        \node[minimum size=0.5cm, fill=red!25] (boxb\i) at (5.5+\i, 0) {\j};
    }
    
    \draw [decorate,decoration={brace,amplitude=5pt,mirror,raise=5pt}] (5.25, -0.25) -- (6.25, -0.25) node [black,midway,yshift=-25pt, align=center] {\footnotesize match final\\[-2pt]\footnotesize$k$ symbols};
    \draw [decorate,decoration={brace,amplitude=5pt,mirror,raise=5pt}] (-0.25, -0.25) -- (0.75, -0.25);
    \draw [decorate,decoration={brace,amplitude=5pt,mirror,raise=5pt}] (2.25, -0.25) -- (3.25, -0.25);
    \draw[-] (4.85, -1.125) -| (0.25, -0.57);
    \draw[-] (4.85, -1.125) -| (2.75, -0.57);

    \draw[->] (1, -0.425) |- (1.35, -1.65);
    \draw[->] (3.5, -0.425) |- (3.15, -1.65);

    \node[align=center] at (2.25,-1.65) {\footnotesize Uniform \\[-2pt] \footnotesize distribution};
\end{tikzpicture}
    \caption{Conditional $k$-gram model. The conditional $k$-gram is the in-context estimate of the Markov process and is realized in two steps. The first step is to find the locations in the sequence (marked red) which match the final $k$ symbols (functionally, a \kth induction head). The conditional $k$-gram model returns the uniform distribution over the next symbol at these locations (marked blue).}
    \label{fig:cond-k-gram}
\end{figure}

There is a large body of active research focused on studying different aspects of transformer models \cite{weiss2021thinking,giannou23looped,oymak2023attn-prompt,li2023theoretical}. Our work closely relates to the aspects of understanding the representation power of transformers, and in-context learning. \cite{Yun2020seq2seq,perez2021turing,wei2022turing-approx} study the representation capabilities of transformers and show properties such as universal approximation and Turing-completeness. Viewing transformers as sequence to sequence models, \cite{liu2023transformers,bhattamishra2020ability} study their ability to model formal languages and automata. Along more related lines to our work, \cite{sanford2024transformers,NEURIPS2023_73bf6924} present logarithmic depth transformer constructions for representing a $k$-hop generalization of the notion of an induction head \cite{olsson2022incontext}. On the other hand the theoretical and mechanistic understanding of in-context learning \cite{wei2022chain} has received much attention lately \cite{bai2023transformers,lin2024dual,akyürek2023learning,hoogland2024developmental}, focusing on different operating regimes and phases of learning. There are a few recent papers which study the behavior of transformers when trained on data generated from Markov processes, and generalizations thereof \cite{bietti2023birth,rajaraman2024theory}. In particular, \cite{makkuva2024attention,nichani2024transformers} study the optimization landscape of gradient descent in learning generalizations of Markov processes, and \cite{edelman2024evolution} present a study of how transformers learn to represent in-context $k$-gram models, focusing on different phases of learning.

\section{Preliminaries}
\label{sec:prelim}

We provide the necessary background for Markov processes, the conditional $k$-gram model, and the transformer architecture.

\subsection{Markov processes}

Markov processes are one of the widely used models in sequence modeling \cite{norris1998markov}. The characterizing property of these processes is that at any time step, the future evolution is only influenced by the most recent states. More formally, a sequence $(X_n)_{n \geq 1}$ is a \kth Markov process on a finite state space $[S]$ with the transition kernel $P$, if surely,
\begin{align*}
 P\big(X_{n+1} \mid X_1,\cdots,X_n \big) &= P\big(X_{n+1} \mid X_{n-k+1},\cdots,X_n \big)
\end{align*}

This property allows us to capture the conditional distribution at any position using only its previous $k$ symbols. This motivates the notion of a conditional $k$-gram, its empirical counterpart, defined for any sequence $(x_1, \ldots, x_n)$.
\begin{definition}[Conditional $k$-gram model] \label{def:condK}
Given a sequence $(x_1,\cdots,x_n)$ of length $n$ in $[S]^n$, the conditional $k$-gram model $\widehat{\operatorname{Pr}}_k(\cdot \mid x_1,\cdots,x_n)$ corresponds to the in-context estimate of the distribution over symbols conditioned on the last $k$ symbols, \ie for $x \in \set S$,
\begin{align}
     &\widehat{\operatorname{Pr}}_k(x \mid x_1,\cdots,x_n) \nonumber\\ &\define \frac{\sum_{i=k+1}^n \mathbb{I} (x_i = x, x_{i-1} = x_n, \ldots, x_{i-k} = x_{n-k+1})} {\sum_{i=k+1}^n \mathbb{I} (x_{i-1} = x_n, \ldots, x_{i-k} = x_{n-k+1})},
     \label{eq:kgram}
\end{align}
\end{definition}
which is defined only so long as the denominator is non-zero. This structure is illustrated in \Cref{fig:cond-k-gram}. It is well known that the conditional $k$-gram in \prettyref{eq:kgram} with Laplace smoothing corresponds to the Bayes optimal estimate of the next symbol probability, when the data is drawn from fixed Markov process sampled from a prior distribution \cite{norris1998markov}.

In our experiments, we will consider \kth Markov kernels sampled from a Dirichlet prior with parameter $\bm{1}$. Namely, the transition $P(\cdot|X_1=i_1,\cdots,X_k=i_k)$ is sampled independently and uniformly on the $S$-dimensional simplex $\Delta_1^S$, for each tuple $(i_1,\cdots,i_k)$.

\subsection{Transformer architecture} 

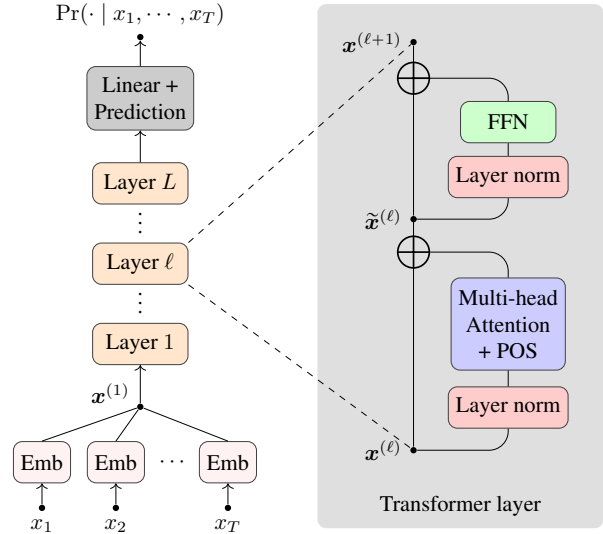
\begin{figure}[t]
\centering
\resizebox{8cm}{!}{
  \begin{tikzpicture}[
    node distance=2cm,
    startstop/.style={rectangle, rounded corners, minimum width=1.5cm, minimum height=0.65cm,text centered, draw=black!75},
    process/.style={rectangle, minimum width=3cm, minimum height=1cm, text centered, draw=black, fill=orange!30},
    io/.style={trapezium, trapezium left angle=70, trapezium right angle=110, minimum width=3cm, minimum height=1cm, text centered, draw=black, fill=blue!30},
    decision/.style={diamond, minimum width=3cm, minimum height=1cm, text centered, draw=black, fill=green!30},
    every text node part/.style={align=center},
    cross/.style={path picture={ 
  \draw[black]
(path picture bounding box.east) -- (path picture bounding box.west) (path picture bounding box.south) -- (path picture bounding box.north);
}}
    ]
    \fill[fill=gray!25, rounded corners] (-3,-1.9) rectangle ++(4.5,8.25);
    \node (tarc) at (-0.75,-1.55) {Transformer layer};
    \node (LN1) [startstop, fill=red!20]      {Layer norm};
    \node (xL) [circle,fill,inner sep=1pt, below left = 0.3cm and 0.5cm of LN1] {} node[left = 0cm of xL] {$\bm{x}^{(\ell)}$};
    \node (ATT) [startstop, fill=blue!20, above = 0.25cm of LN1, minimum height=1.45cm]     {Multi-head \\ Attention\\+ POS};

    \node (RES1) [circle, line width=0.75pt, cross, inner sep=5pt, draw, above = 2.8cm of xL] {};

    \node (yL) [circle,fill,inner sep=1pt, above = 0.2cm of RES1] {}     node[left = 0cm of yL] {$\widetilde{\bm{x}}^{(\ell)}$};

    \node (LN2) [startstop, fill=red!20, above = 1.25cm of ATT] {Layer norm};
    \node (FF) [startstop, fill=green!20, above = 0.2cm of LN2] {FFN};

    \node (RES2) [circle, line width=0.75pt, cross, inner sep=5pt, draw, above = 1.9cm of yL] {};
    \node (zL) [circle,fill,inner sep=1pt, above = 0.25cm of RES2] {} node[left = 0cm of zL] {$\bm{x}^{(\ell+1)}$};
    \node (L1) [startstop, fill=orange!20, above left = 1.3cm and 3.5cm of xL, minimum height=0.65cm] {Layer $1$};
    \node (1dots) [above = 0.03cm of L1] {$\vdots$};
    \node (Ll) [startstop, fill=orange!20, above = 0.6cm of L1, minimum height=0.65cm] {Layer $\ell$};
    \node (Ldots) [above = 0.03cm of Ll] {$\vdots$};
    \node (LL) [startstop, fill=orange!20, above = 0.6cm of Ll, minimum height=0.65cm] {Layer $L$};
    \node (in) [circle, fill,inner sep=1pt, below=0.6cm of L1] {} node[above left=-0.1cm and 0cm of in] {$\bm{x}^{(1)}$};
    \node (pred) [startstop, fill=black!20, above = 0.5cm of LL, minimum height=1cm] {Linear +\\ Prediction};
    \node (out) [circle, fill,inner sep=1pt, above=0.4cm of pred] {} node[above=0cm of out] {$\operatorname{Pr} (\cdot \mid x_1,\cdots,x_T)$};
    \node (emb1) [startstop, fill=pink!20, below left = 0.5cm and 1.05cm of in, minimum width = 0.5cm] {Emb};
    \node (emb2) [startstop, fill=pink!20, right = 0.25cm of emb1, minimum width = 0.5cm] {Emb};
    \node (emb3) [startstop, fill=pink!20, right = 0.85cm of emb2, minimum width = 0.5cm] {Emb};
    \node (dots) [right = 0.1cm of emb2] {$\cdots$};
    \node (x1) [circle, fill,inner sep=1pt, below = 0.33cm of emb1] {} node[below=0cm of x1] {$x_1$};
    \node (x2) [circle, fill,inner sep=1pt, below = 0.33cm of emb2] {} node[below=0cm of x2] {$x_2$};
    \node (xn) [circle, fill,inner sep=1pt, below = 0.33cm of emb3] {} node[below=0cm of xn] {$x_T$};

    \draw[rounded corners=3mm, -] (xL.east) -| (LN1.south);
    \draw[rounded corners=3mm, -] (ATT.north) |- (RES1.east);
    \draw[-] (LN1.north) |- (ATT.south);
    \draw[->] (in) -- (L1);
    \draw[->] (pred) -- (out);
    \draw[->] (LL) -- (pred);
    \draw[->] (x1) -- (emb1);
    \draw[->] (x2) -- (emb2);
    \draw[->] (xn) -- (emb3);
    \draw[-] (emb1.north) -- (in);
    \draw[-] (emb2.north) -- (in);
    \draw[-] (emb3.north) -- (in);
    \draw[dashed] (Ll.south east) -- (xL);
    \draw[dashed] (Ll.north east) -- (zL);
    \draw[-] (xL.north) |- (yL.east);
    \draw[rounded corners=3mm, -] (yL.east) -| (LN2.south);
    \draw[rounded corners=3mm, -] (FF.north) |- (RES2.center);
    \draw[-] (yL.north) |- (zL.south);
    \draw[-] (LN2.north) |- (FF.south);

    (L1.south)

  \end{tikzpicture}
  }
  \caption{Transformer architecture. POS refers to the relative position encodings.}
  
  \label{fig:1}
\end{figure}

In this paper, we will consider variants of the standard transformer architecture in \Cref{fig:1} introduced in \cite{vaswani2017attention}, with the goal to understand the role of depth and the non-linearities in the architecture. The simplest variant removes all the layer normalization and the (non-linear) feedforward layer, and is referred to as an {\it attention-only} transformer. The $L$-layer $1$-head attention-only transformer with relative position encodings, operating on a sequence of length $T$ is mathematically defined in Architecture~\ref{arch:1}.

\begin{Architecture}
\begin{algorithmic}
\FOR{$n=1,2,\cdots,T$}
\STATE \vspace{-1.5em}\begin{flalign}
    \bm{x}_n^{(1)} = \texttt{Emb} (x_n) \in \mathbb{R}^d.&& \tag{Input embeddings}
\end{flalign}
\ENDFOR
\FOR{$\ell=1,2,\cdots,L,$}
\FOR{$n=1,2,\cdots,T,$}
\STATE \vspace{-1.15em}\begin{align}
    &\widetilde{\bm{x}}_n^{(\ell)} = \sum\nolimits_{i \in[n]} \operatorname{att}^{(\ell)}_{n, i} \cdot \bm{W}_V^{(\ell)} \left( \bm{x}_i^{(\ell)} + {\color{blue!50!} \bm{p}^{(\ell),V}_{n-i}} \right) \in \mathbb{R}^d, \hphantom{randomrandom}\tag{Attention}\\
    &\bm{x}_n^{(\ell+1)} = \bm{x}_n^{(\ell)} + \widetilde{\bm{x}}_n^{(\ell)}, \tag{Residual}
\end{align}
\ENDFOR
\ENDFOR
\STATE \begin{flalign}
&\operatorname{logit}_T = \bm{A} \bm{x}_T^{(L+1)} + \bm{b} \quad \in \mathbb{R}^S, && \tag{Linear}\\
&\operatorname{Pr} _{\bm{\theta}} \left( \cdot \mid x_1,\cdots,x_T\right) = f \left(\operatorname{logit}_T \right) && \tag{Prediction}
\end{flalign}
\end{algorithmic}
\caption{Attention-only transformer.}
\label{arch:1}
\end{Architecture}

The attention scores in layer $\ell$, $\{ \operatorname{att}_{n,i}^{(\ell)} i \le n \}$, are computed as $\texttt{Softmax} \big( \big\{ \big\langle  \bm{W}_K^{(\ell)} (\bm{x}_j^{(\ell)} + {\color{blue!60} \bm{p}^{(\ell),K}_{n-j}}), \bm{W}_Q^{(\ell)} \bm{x}_j^{(\ell)} \big\rangle : j \in [n] \big\} \big)$.
The superscript $(\ell)$ indicates the layer index, and the matrices $\bm{W}_K^{(\ell)}$, $\bm{W}_Q^{(\ell)},\bm{W}_V^{(\ell)} \in \mathbb{R}^{d \times d}$ capture the key, query and value matrices in layer $\ell$. Note that the attention-only transformer may include a feedforward layer with linear activations, i.e. a linear transformation. For representation purposes, this linear transformation can be combined with the projection matrix in the attention layer, allowing the feedforward layer to be omitted from the model. In the attention layer, we consider relative position encodings (the terms labeled in blue), which translates the key and value vectors depending on the relative position of the embedded symbol.

The extension to $H$ heads is straightforward, where in each transformer layer there are $H$ attention layers in parallel, resulting in $\bm{y}_n^{(\ell,1)},\cdots,\bm{y}_n^{(\ell,H)} \in \mathbb{R}^d$ for each $n$. These vectors are concatenated and passed through a linear transformation $\bm{W}_O^{(\ell)} : \mathbb{R}^{dH} \to \mathbb{R}^d$ which is the output of the attention layer. Finally, the output of the model after $L$ layers is passed through a linear layer, which projects the $d$-dimensional embeddings back into $\mathbb{R}^S$ and the resulting vector is passed through a non-linearity $f$, usually a softmax, to result in the model's prediction of the next symbol probabilities. The theoretical results in this paper will choose $f = \operatorname{ReLU} (\cdot)$.


\begin{figure}[t]
\centering
\begin{subfigure}
    \centering
    \includegraphics[width=0.825\columnwidth]{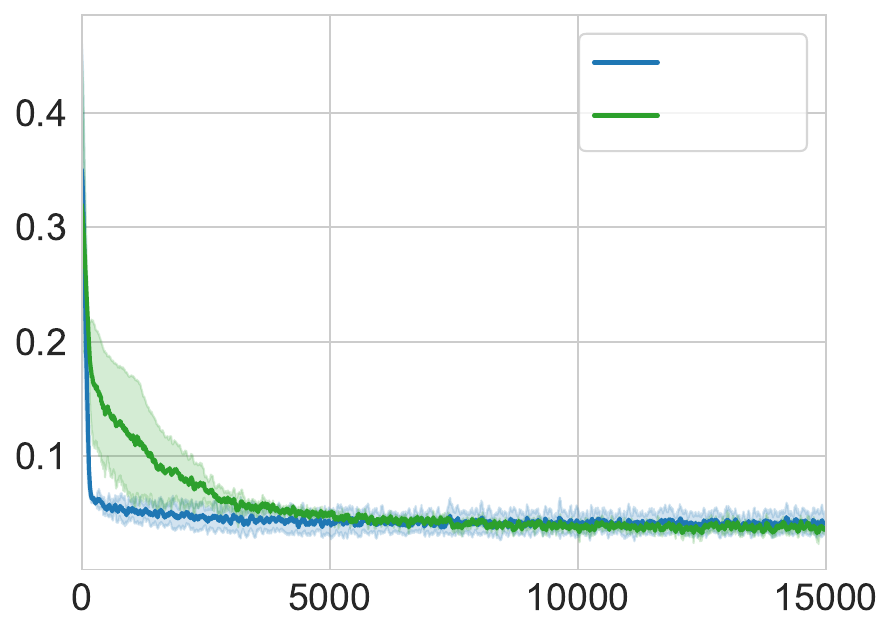}
    \put(-110,-6){\fontsize{8}{3}\selectfont Iteration}
    \put(-202,70){\rotatebox[origin=t]{90}{\fontsize{8}{3}\selectfont Test loss gap from the optimal}}
    \put(-45,122){\fontsize{9}{3}\selectfont $k=2$}
    \put(-45,110){\fontsize{9}{3}\selectfont $k=4$}
\end{subfigure}
\begin{subfigure}
    \centering
    \includegraphics[width=0.83\columnwidth]{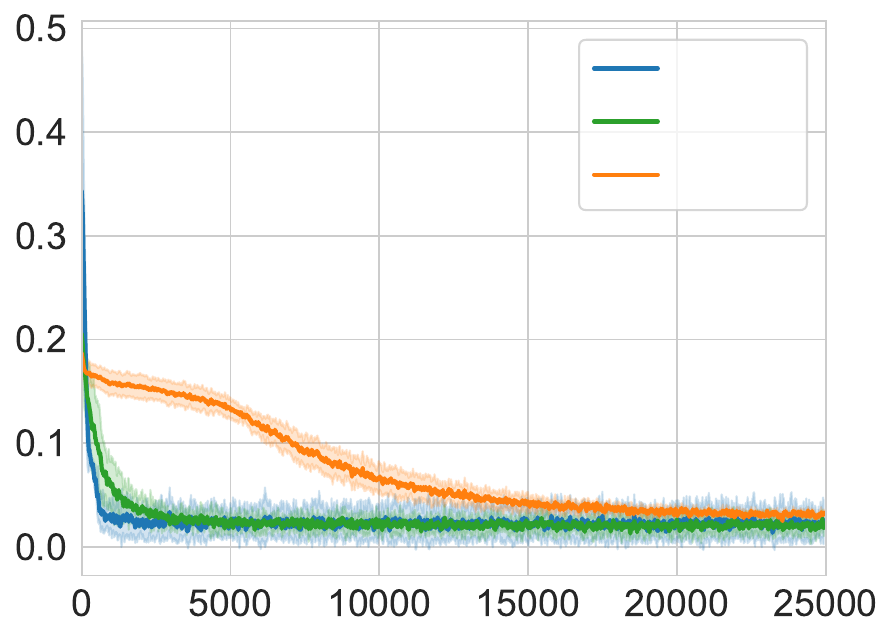}
    \put(-110,-6){\fontsize{8}{3}\selectfont Iteration}
    \put(-202,70){\rotatebox[origin=t]{90}{\fontsize{8}{3}\selectfont Test loss gap from the optimal}}
    \put(-45,123){\fontsize{9}{3}\selectfont $k=2$}
    \put(-45,111){\fontsize{9}{3}\selectfont $k=4$}
    \put(-45,99){\fontsize{9}{3}\selectfont $k=8$}
\end{subfigure}
\caption{Gap with the optimal test loss for $(a)$ a $2$-layer, $1$-head transformer model (above), and $(b)$ a $3$-layer, $1$-head transformer (below), averaged over $3$ runs for each $k$. The models learn the conditional $k$-gram model for randomly sampled $k$-th order Markov processes, even for large $k$.}
\label{fig:order8}
\vspace{-1em}
\end{figure}
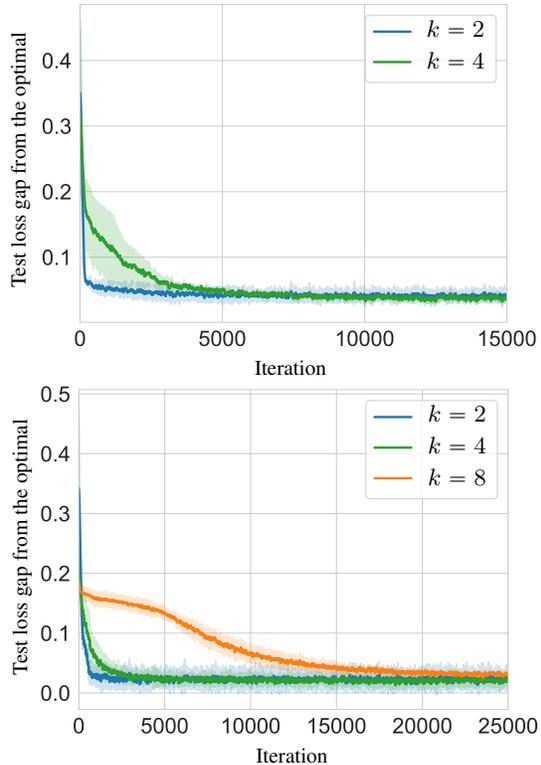

\section{Understanding the empirical behavior of transformers} \label{sec:early-empirical}

The motivation for the present work comes from a series of experimental results, which challenge our current understanding of transformers in the context of learning Markov processes. Several works in the literature \cite{makkuva2024attention,nichani2024transformers,edelman2024evolution} have studied the ability of transformer models to learn \kth Markov processes. The experimental results present in the literature suggest that in order for a $2$ layer transformer model to be able to learn a randomly sampled Markov process of order $k$, it is crucial for the number of heads in the first attention layer to scale linearly with the order, $k$. In particular, the authors of \cite{edelman2024evolution} claim that in their experiments, ``Single attention headed models could not achieve better performance than bigram (models)'' in learning random \kth Markov processes in-context. Similarly, the authors of \cite{nichani2024transformers} study a generalization of learning \kth Markov processes to learning causal processes on degree $k$ graphs. The theory and experiments pertain to $2$-layer transformers with $k$ heads.

In \Cref{fig:order8}, we train $2$ and $3$-layer transformers with a single head on data drawn from random Markov processes of various orders drawn from a Dirichlet prior. With $2$ layers and a single head, we see that the model is able to learn even order-$4$ Markov processes, and go beyond the simple order-$1$ processes which were projected to be the limit of its ability to learn. Likewise, with $3$ layers, transformers are able to go much further and learn order-$8$ Markov processes, which was the largest value of $k$ we evaluated on.

These results contrast with our current understanding of how induction heads are realized in the parameter space \cite{edelman2024evolution,nichani2024transformers} - existing constructions which realize these attention patterns require $k$ heads when the number of layers is $2$, and it's unclear how to implement these them with fewer heads. At a high level, each of the $k$ heads play a critical role - where, loosely speaking, the $i^{\text{th}}$-head looks back $i$ positions in the sequence.

Building up to our main results, in the sequel, we study the simpler case of attention-only transformers where the feedforward layers and layer normalization are removed.  



\section{Warming up: Attention-only transformers} \label{sec:logk}

The study of attention-only transformers trained on Markov processes has garnered some attention in the prior literature. Notably, the authors of \cite{edelman2024evolution} study $2$-layer $1$-head attention-only transformers trained on data drawn from $1^{\text{st}}$-order Markov processes whose parameters are drawn from a Dirichlet prior. The model is observed to learn a very specific behavior, known as an ``induction head'' \cite{olsson2022incontext}, which in this setting is able to represent the conditional $1$-gram (\prettyref{eq:kgram}).

The  The induction head mechanism is composed of two layers where the first layer learns the attention pattern $\operatorname{att}^{(1)}_{n,i} = \mathbb{I} (i=n-1)$, thereby allowing the model to capture information about the symbol at position $n-1$ in the embedding vector at time $n$. In the second layer, the attention layer picks out those indices $n$ where $x_{n-1} = x_T$, the final symbol in the sequence. At these positions, since $x_{n-1} = x_T$, one would expect that the next symbol $x_n$ is a good predictor of $x_{T+1}$, and the model uses this information to predict the next symbol $x_{T+1}$ according to its conditional empirical estimate, $\widehat{\operatorname{Pr}}_1 (x_{T+1} | x_1,\cdots,x_T)$, i.e. the conditional $1$-gram model.


\begin{theorem} \label{theorem:simple}
The conditional $1$-gram model can be represented by a $2$-layer and $1$-head attention-only transformer with embedding dimension $d = 3S+2$.
\end{theorem}

Although a version of this result is also proved in \cite{edelman2024evolution}, we include a proof in \Cref{app:simple} for completeness.

\begin{remark}
In \Cref{theorem:simple} and other results to follow, we de-emphasize the role of the bit-precision to which the transformer is implemented. That said, note that when the constructions in \Cref{theorem:simple,theorem:simple-k,theorem:logk} are implemented to $O(\log(T))$ bits of precision, the representation results are realized up to an additive $O(1/T)$ error.
\end{remark}

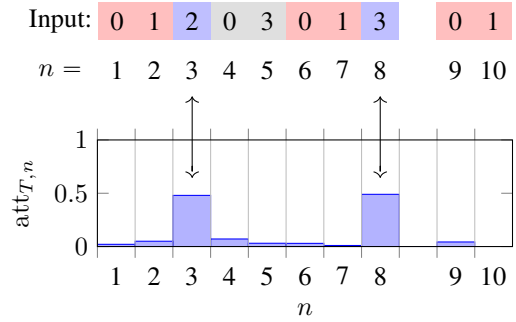
\begin{figure}
    \centering
    \begin{tikzpicture}
    \node[align=left] (ins) at (-0.75,0) {Input:};
    \node at (-0.75, -0.65) {$n=$};
    \foreach \i/\j in {0/0, 0.5/1, 2.5/0, 3/1} {
        \node[minimum size=0.5cm, fill=red!25] (box\i) at (\i, 0) {\j};
    }
    \foreach \i/\j in {1.5/0, 2/3} {
        \node[minimum size=0.5cm, fill=gray!25] (box\i) at (\i, 0) {\j};
    }
    \foreach \i/\j in {1/2, 3.5/3} {
        \node[minimum size=0.5cm, fill=blue!25] (box\i) at (\i, 0) {\j};
    }
    \foreach \i/\j in {0/1, 0.5/2, 1/3, 2.5/6, 1.5/4, 2/5, 3/7, 3.5/8, 4.5/9, 5/10} {
        \node[minimum size=0.5cm] (box\i) at (\i, -0.65) {\j};
    }
    \node[minimum size = 0.5cm] (boxb10) at (5, -3.4) {10};

    \begin{axis}[
        ymin=0, ymax=1,
        xmin=1, xmax=12,
        xtick={1,2,3,4,5,6,7,8,9,10,11,12,13},
        xticklabels={1,2,3,4,5,6,7,8,,9,10,11,12},
        ybar interval,
        area style,
        ytick distance=0.5,
        xshift=-0.26cm,yshift=-3cm,
        width=7.1cm,
        height=3cm,
        ylabel near ticks,
        xlabel near ticks,
        xlabel={$n$},
        ylabel={$\operatorname{att}_{T,n}$}
        ]
        \addplot+[ybar interval,mark=no] plot coordinates { (1, 0.021) (2, 0.05) (3, 0.48) (4, 0.071) (5, 0.031) (6, 0.03) (7, 0.01) (8,0.49) (9,0) (10,0.043) (11,0.0) };
    \end{axis}
    \draw[<-<] (1,-1) |- (1,-2);
    \draw[<-<] (3.5,-1) |- (3.5,-2);

    \foreach \i/\j in {0/0, 0.5/1} {
        \node[minimum size=0.5cm, fill=red!25] (boxb\i) at (4.5+\i, 0) {\j};
    }
\end{tikzpicture}
    \caption{\kth induction head for $k=2$. The attention pattern $\operatorname{att}_{T,n}$ is maximized for those values of $n$ at which $x_{T-j+1} = x_{n-j}$ for all $j \in [k]$. These are the positions where the $k$-length prefix at those positions matches with the last $k$ symbols in the sequence.}
    \label{fig:k-iH}
\end{figure}

The ideas in \Cref{theorem:simple} readily extend to representing the conditional $k$-gram model, by instead using $k$ heads in the first layer. The $j^{\text{th}}$ head learns the attention pattern $\operatorname{att}_{n,i}^{(1)} = \mathbb{I} (i=n-j)$ and concatenating the outputs of the heads, the model learns to aggregate information about $x_n,\cdots,x_{n-k}$ in the embedding vector at time $n$. The second layer realizes what is best described as a ``\kth'' induction head, where the model learns to pick out those positions $n$ where for every $j \in [k]$, $x_{n-j} = x_{T-j+1}$, i.e. the history of length $k$ at those positions match the final $k$ symbols in the input sequence ( see \Cref{fig:k-iH}). This mechanism is also referred to as a long-prefix induction head in the literature \cite{goldowsky2023localizing}.

\begin{definition}[Higher-order induction head] \label{def:k-order_IH}
A $1$-head attention layer is said to realize a \kth induction head if on any sequence $(x_1,\cdots,x_T) \in [S]^T$, for any fixed $n \le T$, as a function of the input sequence, $\operatorname{att}_{n,T}$ is maximized if and only if $x_{n-j} = x_{T-j+1}$ for every $j \in [k]$.
\end{definition}

The \kth induction head generalizes the concept of an induction head \cite{olsson2022incontext}, and is able to keep track of the positions $i \le n$ where there is a perfect occurrence of the final $k$ symbols in the sequence. Such attention patterns are immediately useful in representing the conditional $k$-gram - increasing the temperature within the softmax of this attention layer results in an attention pattern which converges to the uniform distribution over those positions where the final $k$ symbols $x_{T-k+1},\cdots,x_T$ are seen previously in the sequence. Loosely, this is what allows the model to ``condition'' on the last $k$ symbols in the sequence. With $k$ heads, the model can aggregate information from the previous $k$ positions and implement a \kth induction head, which leads to the following result. A full proof is discussed in \Cref{app:simple-k}.

\begin{theorem} \label{theorem:simple-k}
The conditional $k$-gram model can be represented by an attention-only transformer with $2$ layers, $k$ heads and embedding dimension $d = (k+2)S+k+1$.
\end{theorem}

While this result is positive, it suggests that a $2$-layer transformer requires approximately $k$ times as many parameters to be able to represent the conditional $k$-gram model. The first result we prove is that increasing the depth of the model is exponentially more beneficial, in that a transformer with $O(\log (k))$ depth can estimate in-context $k$-grams.

\begin{theorem} \label{theorem:logk}
The conditional $k$-gram model can be represented by an attention-only transformer with relative position encodings, with $L = \lceil \log_2 (k+1) \rceil$ layers and $1$ head per layer. The embedding dimension is $\le 2k (S+1) + S$.
\end{theorem}

With $2$ layers and $k$ heads, the transformer aggregates information about each of the previous $k$ positions one step at a time through the $k$ heads. However, with $\Omega(\log (k))$ layers, the same task can be done far more efficiently. In the first attention layer, the model aggregates information about the current and previous position. Namely, using the relative position embeddings, $\bm{x}_n^{(2)}$ is chosen as a linear combination of $\bm{x}_n^{(1)} = \texttt{Emb} (x_n)$ and $\bm{x}_{n-1}^{(1)} = \texttt{Emb} (x_{n-1})$. This allows the embedding at position $n$ to aggregate information about $x_n$ and $x_{n-1}$. In the same vein, in the second attention layer, the model aggregates information from $\bm{x}_n^{(2)}$ and $\bm{x}_{n-2}^{(2)}$ in $\bm{x}_n^{(3)}$; the former has information about $x_n$ and $x_{n-1}$, and the latter has information about $x_{n-2}$ and $x_{n-3}$. This expands the ``window'' of $x_i$'s on which $\bm{x}_n$ depends on to size $4$. In the $\ell^{\text{th}}$ layer, the model aggregates information from $\bm{x}_n^{(\ell)}$ and $\bm{x}_{n-2^\ell}^{(\ell)}$ which allows $\bm{x}_n^{(\ell+1)}$ to effectively depend on the $x_i$'s in a window of size $2^{\ell+1}$ starting at position $n$, namely $x_n,\cdots,x_{n-2^{\ell+1}+1}$. In the final layer, the embedding at position $i$, $\bm{x}_i^{(L)}$ for $L = \lceil \log_2(k+1) \rceil$ depends on $x_n, x_{n-1},\cdots,x_{n-k}$. In the last layer, the model can realize the dot-product $\left\langle \bm{W}_K^{(L)} \bm{x}_n^{(L)}, \bm{W}_Q^{(L)} \bm{x}_T^{(L)} \right\rangle = \sum_{j=1}^k \mathbb{I} ( x_{n-j} = x_{T-j+1} )$ by choosing the key and query vectors appropriately. By increasing the temperature in the attention softmax, the attention pattern realized is the uniform distribution on values of $n$ such that $x_{n-j} = x_{T-j+1}$ for every $j \in [k]$, i.e., a \kth induction head. The full proof of this result is provided in \Cref{app:logk}.

While this is a promising step toward understanding the behavior transformers exhibit in \Cref{fig:order8}, showing that depth plays an important role in their ability to represent conditional $k$-gram models, the picture is still not complete. The experimental results in \Cref{sec:early-empirical} do not preclude the possibility that a transformer might not even require logarithmic depth to be able to learn \kth Markov processes approximately. In the next section, we will study constant-depth transformers and establish a rather surprising positive result about the representation power of this class of models in capturing conditional $k$-grams.

\input{constant-depth}

\newpage
\bibliographystyle{plainnat}
\bibliography{main}
\newpage
\input{appendix}




\end{document}

%% file: commands_icml.tex
\usepackage[utf8]{inputenc} 
\usepackage[T1]{fontenc}    
\usepackage{url}            
\usepackage{booktabs}       
\usepackage{amsfonts}       
\usepackage{nicefrac}       
\usepackage{microtype}      
\usepackage{tabu}
\usepackage{multicol}
\usepackage{soul}
\usepackage{bbm}
\usepackage{lipsum}
\usepackage{kantlipsum}
\usepackage{tabularx}

\usepackage{cancel}

\usepackage{amsmath,amssymb,amsfonts,amsthm, dsfont, color}
\usepackage{mathtools}
\usepackage{graphicx}
\usepackage{textcomp}
\usepackage{xcolor, fancyhdr}
\usepackage{enumitem}
\usepackage{float}
\usepackage{nicefrac}

\usepackage{wrapfig}
\usepackage{mathtools}
\usepackage{cuted}

\definecolor{MyGreen1}{RGB}{20,180,40}
\usepackage{multirow, tikz}
\allowdisplaybreaks

\usepackage{nicefrac,color,mathrsfs}
\usepackage{multirow,caption,tikz,subcaption}
\usetikzlibrary{shapes.misc, positioning}
\usetikzlibrary{decorations.pathreplacing}
\usetikzlibrary{arrows.meta, shapes,patterns.meta}

\tikzset{
  block/.style    = {draw, thick, rectangle, minimum width = 2em},
sblock/.style      = {draw, thick, rectangle, minimum height = 2em,
minimum width = 2em}, 
}

\usepackage{comment}

\newcommand{\gptwo}{\textsc{GPT-2}\xspace}

\newcommand{\set}[1]{[#1]}

\DeclarePairedDelimiterX{\infdivx}[2]{(}{)}{%
  #1\;\delimsize\|\;#2%
}

\usepackage{thmtools}

\newtheorem{claim}{Claim}

\usepackage{mathabx}

\usepackage{prettyref,xspace}
\usepackage{tikz}

\newrefformat{cond}{Condition~\ref{#1}}
\newrefformat{eq}{Eq.~\eqref{#1}}
\newrefformat{thm}{Thm.~\ref{#1}}
\newrefformat{th}{Theorem~\ref{#1}}
\newrefformat{chap}{Chapter~\ref{#1}}
\newrefformat{sec}{Sec.~\ref{#1}}
\newrefformat{algo}{Algorithm~\ref{#1}}
\newrefformat{fig}{Fig.~\ref{#1}}
\newrefformat{tab}{Table~\ref{#1}}
\newrefformat{rmk}{Remark~\ref{#1}}
\newrefformat{clm}{Claim~\ref{#1}}
\newrefformat{def}{Definition~\ref{#1}}
\newrefformat{cor}{Corollary~\ref{#1}}
\newrefformat{lmm}{Lemma~\ref{#1}}
\newrefformat{prop}{Proposition~\ref{#1}}
\newrefformat{pr}{Proposition~\ref{#1}}
\newrefformat{app}{App.~\ref{#1}}
\newrefformat{prob}{Problem~\ref{#1}}
\newrefformat{ques}{Question~\ref{#1}}
\newrefformat{notee}{Note~\ref{#1}}
\newrefformat{assump}{Assumption~\ref{#1}}
\newrefformat{issuee}{Issue ~\ref{#1}}
\newrefformat{fix}{Fix ~\ref{#1}}

\newcommand{\naturals}{\mathbb{N}}

\newcommand{\bmhat}[1]{\widehat{\bm{#1}}}

\newcommand{\vect}[1]{\boldsymbol{#1}}

\newcommand{\bx}{\vect{x}}
\newcommand{\by}{\vect{y}}

\newcommand{\bM}{\vect{M}}

\usepackage{derivative}

\newcommand{\define}{\triangleq}

\newcommand{\ie}{i.e.\xspace}

\mathchardef\mhyphen="2D

\definecolor{MyGreen1}{RGB}{20,180,40}

\usepackage{siunitx}

\newcommand{\kth}{$k^{\text{th}}$-order~}

\usepackage{minitoc}
\usepackage{etoc}

%% file: constant-depth.tex
\section{Understanding the role of non-linearity: Constant-depth constructions}
\label{sec:constant}
In the previous section, we saw how the transformer uses the power of depth to learn conditional $k$-grams far more efficiently. In particular, every additional attention layer effectively doubles the window of positions $i=n-1,n-2,\cdots$ which the model has access to information about at the current time $n$. By composing $L = \Omega (\log(k))$ attention layers, the model is able to collect enough information within the output embedding $\bm{x}_n^{(L+1)}$ to be able to realize a $k^{\text{th}}$-order induction head in the next layer. In this section, we prove that adding non-linearity to the architecture, in the form of layer normalization, can significantly change the mechanism in which the transformer realizes this $k^{\text{th}}$-order head. In particular, there are constant depth architectures which allow a $k^{\text{th}}$-order induction head to be realized, surpassing the logarithmic depth attention-only constructions.

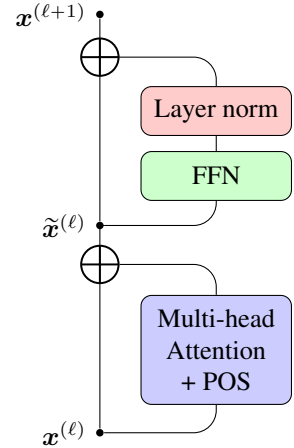
\begin{figure}[t]
\centering
  \begin{tikzpicture}[
    node distance=2cm,
    startstop/.style={rectangle, rounded corners, minimum width=2cm, minimum height=0.75cm,text centered, draw=black},
    process/.style={rectangle, minimum width=3cm, minimum height=1cm, text centered, draw=black, fill=orange!30},
    io/.style={trapezium, trapezium left angle=70, trapezium right angle=110, minimum width=3cm, minimum height=1cm, text centered, draw=black, fill=blue!30},
    decision/.style={diamond, minimum width=3cm, minimum height=1cm, text centered, draw=black, fill=green!30},
    every text node part/.style={align=center},
    cross/.style={path picture={ 
  \draw[black]
(path picture bounding box.east) -- (path picture bounding box.west) (path picture bounding box.south) -- (path picture bounding box.north);
}}
    ]

    \node (ATT) [startstop, fill=blue!20, minimum height=1.45cm]     {Multi-head \\ Attention\\+ POS};
    \node (xL) [circle,fill,inner sep=1pt, below left = 0.33cm and 0.5cm of ATT] {} node[left = 0cm of xL] {$\bm{x}^{(\ell)}$};

    \node (RES1) [circle, line width=0.75pt, cross, inner sep=5pt, draw, above left = 0.22cm and 0.355cm of ATT] {};

    \node (yL) [circle,fill,inner sep=1pt, above = 0.2cm of RES1] {} node[left = 0cm of yL] {$\widetilde{\bm{x}}^{(\ell)}$};

    \node (FF) [startstop, fill=green!20, above = 1.25cm of ATT, minimum height=0.65cm] {FFN};
    \node (LN2) [startstop, fill=red!20, above = 0.2cm of FF, minimum height=0.65cm] {Layer norm};

    \node (RES2) [circle, line width=0.75pt, cross, inner sep=5pt, draw, above left = 0.2cm and 0.355cm of LN2] {};
    \node (zL) [circle,fill,inner sep=1pt, above = 0.25cm of RES2] {} node[left = 0cm of zL] {$\bm{x}^{(\ell+1)}$};

    \draw[rounded corners=3mm, -] (xL.east) -| (ATT.south);
    \draw[rounded corners=3mm, -] (ATT.north) |- (RES1.east);
    \draw[-] (xL.north) |- (yL.east);
    \draw[rounded corners=3mm, -] (yL.east) -| (FF.south);
    \draw[rounded corners=3mm, -] (LN2.north) |- (RES2.center);
    \draw[-] (yL.north) |- (zL.south);
    \draw[-] (FF.north) |- (LN2.south);

  \end{tikzpicture}
  \caption{Rearranged transformer layer with layer normalization and FFN.}
  \label{fig:3a}
\end{figure}

\paragraph{Modification to the standard transformer architecture.} To simplify the proof of our main result, we will consider a subtle modification to the standard transformer architecture, which is presented in Architecture~\ref{arch:2} and \Cref{fig:3a}. We will remove the first layer norm prior to the multi-head attention and move the second layer norm to after the feed-forward network. It is important to note that \Cref{theorem:main} holds even for the architecture presented in \Cref{fig:1}, which is the architecture we evaluate empirically. The modification we present in \Cref{fig:3a} allows the construction to be simpler and makes it much easier to convey the key intuition. The main difference compared to the attention-only design presented in Architecture~\ref{arch:1} is the addition of layer normalization and a feedforward layer in the for-loop over $n \in [T]$ for each transformer layer $\ell$. The differences between Architectures~\ref{arch:2} and \ref{arch:1} are emphasized in blue. 

\begin{Architecture}
\begin{algorithmic}
\STATE \vspace{-1.15em}\begin{flalign}
    &\quad
    \widetilde{\bm{x}}_n^{(\ell)} = \bm{x}_n^{(\ell)} + \sum\nolimits_{i \in[n]} \operatorname{att}^{(\ell)}_{n, i} \cdot \bm{W}_V^{(\ell)} \left( \bm{x}_i^{(\ell)} + \bm{p}^{(\ell),V}_{n-i} \right),&& \tag{Attention + Residual$_1$}\\
&\quad{\color{blue} \bm{y}_n^{(\ell)} = \bm{W}_2^{(\ell)} \operatorname{ReLU}\left(\bm{W}_1^{(\ell)} \widetilde{\bm{x}}_n^{(\ell)} \right) \in \mathbb{R}^d},&& \tag{FFN} \\
&\quad{\color{blue} \bm{l}_n^{(\ell)} = \frac{\bm{y}_n^{(\ell)} - \mu \bm{1}_{d \times 1}}{\sigma} \in \mathbb{R}^d},&& \label{eq:LN}\tag{LN} \\
&\quad\bm{x}_n^{(\ell+1)} = \bm{l}_n^{(\ell)} + \widetilde{\bm{x}}_n^{(\ell)} \in \mathbb{R}^d,&& \tag{Residual$_2$}
\end{flalign}
\end{algorithmic}
\caption{Modified transformer architecture. The computations above are carried out for each $n \in [T]$ in each layer $\ell \in [L]$. In the layer normalization step (\ref{eq:LN}), the feature mean $\mu$ is defined as, $\mathbb{E}_{i \sim \operatorname{Unif} ([d])} \big[ \big\langle e_i^d \bm{y}_n^{(\ell)} \big\rangle \big]$ and the feature variance $\sigma^2 = \mathbb{E}_{i \sim \operatorname{Unif} ([d])} \big[ \big\langle e_i^d, \bm{y}_n^{(\ell)} \big\rangle^2 \big] - \mu^2$.}
\label{arch:2}
\vspace{-1em}
\end{Architecture}

\begin{figure}[t]
\centering
\begin{tikzpicture}[
    node distance=2cm,
    startstop/.style={rectangle, rounded corners, minimum width=1.65cm, minimum height=0.65cm,text centered, draw=black},
    process/.style={rectangle, minimum width=3cm, minimum height=1cm, text centered, draw=black, fill=orange!30},
    io/.style={trapezium, trapezium left angle=70, trapezium right angle=110, minimum width=3cm, minimum height=1cm, text centered, draw=black, fill=blue!30},
    decision/.style={diamond, minimum width=3cm, minimum height=1cm, text centered, draw=black, fill=green!30},
    every text node part/.style={align=center},
    cross/.style={path picture={ 
  \draw[black]
(path picture bounding box.east) -- (path picture bounding box.west) (path picture bounding box.south) -- (path picture bounding box.north);
}}
    ]

    \node (l1) [startstop, fill=orange!20,  minimum width=2.25cm]      {Layers $1+2$};
    \node (l2out) [circle,fill,inner sep=1pt, above = 0.45cm of l1] {}; \node (l2outlabel) [left = 1.7cm of l2out] {$\arrayrulecolor{gray}\left[\begin{array}{@{}c|c|c|c@{}}
& \texttt{Emb} (x_n) & & \texttt{Emb} (x_T) \\[6pt] \ \cdots & \frac{\bm{u}_{n}}{\| \bm{u}_{n} \|_2} & \cdots & \frac{\bm{u}_{T}}{\| \bm{u}_{T} \|_2} \\[6pt] & \frac{\bm{v}_{n}}{\| \bm{v}_{n} \|_2} & & \frac{\bm{v}_T}{\| \bm{v}_T \|_2}
\end{array}\right]$};
    \node (l3) [startstop, fill=red!20, above = 1cm of l1]      {Layer $3$};
    \node (l1in) [circle,fill,inner sep=1pt, below = 0.3cm of l1]  {};
    \node (l1inlabel) [left = 1.82cm of l1in] {$\arrayrulecolor{gray}\left[\begin{array}{@{}c|c|c|c@{}}
\ \cdots & \texttt{Emb} (x_n) & \cdots & \texttt{Emb} (x_T) \ \
\end{array}\right]$};
    \node (l3out) [circle, fill, inner sep=1pt, above = 0.65cm of l3] {};
    \node (l3outlabel) [left = 1.95cm of l3out] {$\operatorname{att}_{T,n} \propto \exp \left( \kappa \frac{\langle \bm{v}_n, \bm{u}_T \rangle}{\| \bm{v}_n \|_2 \| \bm{u}_T \|_2} \right)$ \\ {\small(realizes a $k^{\text{th}}$-order induction head)}};
    \node (emb) [startstop, below = 0.25cm of l1in, fill=pink!20] {$\texttt{Emb}$};
    \node (l0in) [circle,fill,inner sep=1pt, below = 0.35cm of emb] {} node [below = 0cm of l0in] {$x_1,\cdots,x_T$};

    \draw[-,color=gray] (l1in.west) -| (l1inlabel.east);
    \draw[-,color=gray] (l2out.west) -| (l2outlabel.east);
    \draw[-,color=gray] (l3out.west) -| (l3outlabel.east);
    \draw[-] (l1in.south) -| (l1.south);
    \draw[-] (emb.north) -| (l1in.south);
    \draw[-] (l1.north) -| (l3.south);
    \draw[-] (l3.north) -| (l3out.south);
    \draw[-] (l0in.south) -| (emb.south);
  \end{tikzpicture}
\caption{\textit{Disassembling the constant-depth construction.} The first two layers are critical in the model's ability to capture information from the previous $k$ positions. Layer normalization plays a critical role in in the $3^{\text{rd}}$ layer which realizes a $k^{\text{th}}$-order induction head.}
\label{fig:3b}
\vspace{-2em}
\end{figure}
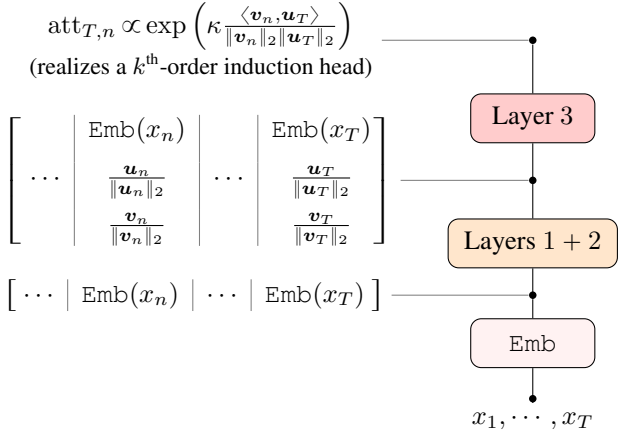


\begin{theorem} \label{theorem:main}
Conditional $k$-grams can be represented by a transformer with $3$ layers, $1$ head per layer, relative position encodings and layer normalization. The embedding dimension is $O(S)$.
\end{theorem}

\begin{remark}
Although the proof stated does not bound the approximation error arising from a finite bound on the bit precision of the transformer, in theory, it should suffice to have $\Omega (\log(T) + k)$ bits per parameter for the statement of \Cref{theorem:main} to go through with an $O(1/T)$ additive approximation error. The main point is that none of the weights of the model exceed $\exp(k)$ and with $\log(T)$ additional bits per parameter, the approximation error scales as $O(1/T)$.
\end{remark}


\subsection{Proof sketch} In the attention-only transformer with $2$ layers and $k$ heads, the model is able to keep track of where the final $k$ symbols in the sequence appeared previously (i.e., a \kth induction head) by, loosely, using each head to keep track of the occurrences of one of the final $k$ symbols. On the other hand, with the benefit of more depth, with $L = \Omega (\log(k))$ layers, the model is able to collect enough information within the output embedding $\bm{x}_n^{(L+1)}$ to be able to realize the same behavior. However, neither of these constructions scale down to the case when the depth and number of heads of the transformer are both constants independent of $k$. We provide a brief intuition for the construction below.

Recall that a \kth induction head keeps track of the indices $i$ such that $\forall j \in [k],\ x_{i-j} = x_{n-j+1}$. Defining $\bm{z}_i \triangleq \sum_{j=1}^k 2^j e_{x_{i-j+1}}$, notice that the condition $\{ \forall j \in [k],\ x_{i-j} = x_{n-j+1}\}$ can equivalently be captured by writing $\{ \bm{z}_{i-1} = \bm{z}_{n} \}$. This true because of the fact that the binary representation of any integer is unique. Furthermore, these vectors, up to scaling, can be realized by softmax attention (namely, $\operatorname{att}_{n,n-i} \propto \ 2^{i}$ for $1 \le i \le k$).

With this step, finding occurrences of the last $k$ symbols in the input sequence boils down to realizing an attention pattern in the second layer, $\operatorname{att}^{(2)}_{n,i}$, which is maximized whenever $\bm{z}_{i-1} = \bm{z}_{n}$. While dot-product attention naively encourages those values of $i$ for which $\bm{z}_{i-1}$ and $\bm{z}_n$ are ``similar'' to each other, a qualitative statement is lacking. In general, it will turn out to that a different measure of similarity is necessary within the softmax to be able to encourage those values of $i$ for which these vectors match. This is where the role of layer-normalization comes in.

Instead of the usual dot-product, suppose the attention mechanism in the second layer was,
\begin{align} \label{eq:L2att}
    \operatorname{att}_{n,i}^{(2)} \propto \exp \left( - \kappa \left\| \frac{\bm{z}_{i-1}}{\| \bm{z}_{i-1} \|_2} - \frac{\bm{z}_n}{\| \bm{z}_n \|_2} \right\|_2^2 \right),
\end{align}
where $\kappa$ is the temperature parameter. Then, as the temperature $\kappa$ grows, the attention pattern essentially focuses on those values of $i$ for which $\bm{z}_i / \| \bm{z}_{i-1} \|_2 = \bm{z}_n / \| \bm{z}_n \|_2$. With this attention pattern, we are thus very close to the statement we wanted to check, ($\bm{z}_{i-1} \overset{?}{=} \bm{z}_n$). As it turns out, for the special structure in the $\bm{z}_i$'s considered (dyadic sums of one-hot vectors), we may write down,
\begin{align*}
    \bm{z}_{i-1} = \bm{z}_n \iff \bm{z}_{i-1} / \| \bm{z}_{i-1} \|_2 = \bm{z}_n / \| \bm{z}_n \|_2.
\end{align*}
A quantifiable equivalence is provided in \Cref{lemma:normerror}.


\paragraph{Realizing $L_2$-norm attention (eq.~\eqref{eq:L2att}).}
Observe the equivalence,
\begin{align}
    \left\langle \frac{\bm{z}_{i-1}}{\| \bm{z}_{i-1} \|_2}, \frac{\bm{z}_n}{\| \bm{z}_n \|_2} \right\rangle = 1 - \frac{1}{2} \left\| \frac{\bm{z}_{i-1}}{\| \bm{z}_{i-1} \|_2} - \frac{\bm{z}_n}{\| \bm{z}_n \|_2} \right\|_2^2
\end{align}
Taking a softmax on both sides, notice that the RHS (up to an additive constant) is the $L_2$-norm based attention, while the LHS is the usual dot-product attention between $\bm{z}_{i-1} / \| \bm{z}_{i-1} \|_2$ and $\bm{z}_n / \| \bm{z}_n \|_2$. Thus on unit-normalized vectors, the $L_2$-norm attention and scaled-dot product attention are nothing but the same.

While the first layer of the transformer computes the $\bm{z}_i$'s by a weighted summation, layer normalization fills in the last missing piece of the puzzle which is to normalize them to unit norm. This is a consequence of defining the embedding vectors appropriately, in such a way that the feature variance evaluates to $\| \bm{z}_i \|_2^2$.

From this step, realizing the actual conditional $k$-gram model follows readily. In particular, as the temperature $\kappa$ in the attention grows, the attention pattern zooms in on indices $i \in \mathcal{I}_n \triangleq \{ k+1 \le i \le n : \forall j \in [k], x_{i-j} = x_{n-j+1} \}$ in the last layer. The value vectors at this step are the one-hot encoding of $x_i$; putting everything together, the logits realized by the transformer are,
\begin{align} \label{eq:logit}
    \operatorname{logit}_T (x_{T+1}) = \frac{1}{|\mathcal{I}_n|} \sum_{i \in \mathcal{I}_n} \mathbb{I} ( x_i = x_T),
\end{align}
which is the conditional $k$-gram model (eq.~\eqref{eq:kgram}).

While the transformer construction described above only requires two layers, the actual construction we propose differs slightly and has an additional layer. The first two layers of the transformer respectively compute $\bm{z}_i$ and $\bm{z}_{i-1}$ which are added to the embedding vector at time $i$. This is important because we need to test whether $\bm{z}_{i-1} \overset{?}{=} \bm{z}_n$ and not whether $\bm{z}_i \overset{?}{=} \bm{z}_n$ or $\bm{z}_{i-1} \overset{?}{=} \bm{z}_n$.

\textbf{Summary.} The construction can be summarized as follows: the first layer computes $\bm{z}_n = \sum_{j=1}^{k} 2^{j-1} \cdot e_{x_{n-j}}$ by choosing appropriate value vectors and relative position embeddings to realize the attention pattern $\operatorname{attn}_{n,n-i} \propto \ 2^i \mathbb{I} (1 \le i \le k)$. The layer normalization that follows subsequently can be made to carry out $L_2$ normalization, by a simple trick which we discuss in \Cref{sec:trick}, resulting in $\bm{z}_n / \| \bm{z}_n \|_2 $ to be appended to the embedding at time $n$. Using a very similar construction, layer $2$ computes $\bm{z}_{n-1} / \| \bm{z}_{n-1} \|_2$, which is added to the embedding at time $n$. Finally, in the last layer, the dot-product $\left\langle \frac{\bm{z}_{i-1}}{\| \bm{z}_{i-1} \|_2}, \frac{\bm{z}_n}{\| \bm{z}_n \|_2} \right\rangle$ defines the attention score, and as the temperature $\kappa$ grows, the pattern converges to $\operatorname{Unif} (\mathcal{I}_n)$. Choosing the value vectors in this layer appropriately results in eq.~\eqref{eq:logit}.

\section{Lower bounds on transformer size}

Having established positive results on the representation power of the standard transformer model in the previous section, we ask what are the limits of how shallow the model can be made to be able to capture conditional $k$-grams. The first result we establish in this vein is a lower bound against $1$ layer transformers showing that their expressive power is too limited unless the embedding dimension or number of heads scale near-linearly in $T$.

\begin{theorem} \label{theorem:lb-1layer}
Consider any $1$-layer transformer with layer normalization and feedforward layers, where all the coordinates of the embedding vectors and unnormalized attention scores are computed with $p$ bits of precision. If the transformer is able to compute the conditional $3$-gram on inputs drawn from $\{ 0,1,2 \}^T$ to within an additive error of $1/3T$, then $2 p H + d p + 2 \ge T/3$.
\end{theorem}

Choosing the bit precision to be $p = O(\log (T))$, this implies that for transformers with $1$ layer, the sum of the number of heads and the embedding dimension must be at least $\Omega (T/\log(T))$, in order to represent conditional $3$-grams to within an additive error of $1/3T$. 

\subsection{Conditional lower bounds on attention-only transformers} \label{subsec:cond}

While the previous section shows that $1$-layer transformers have fairly limited representation power, it is not immediately clear how whether any of these issues are present with transformers with more layers. Indeed, as we discussed in \cref{sec:logk}, an attention-only transformer with $O (\log_2 (k))$ layers and $1$ head per layer can represent conditional $k$-grams on its input sequences. With the addition of non-linearities, \Cref{theorem:main} shows that the model can represent conditional $k$-grams using just a constant number of layers. In this section, we try to understand the gap between these two results and prove conditional lower bounds on the size of attention-only transformers which do not have non-linearities arising from layer normalization. This sheds some more light on the components of the architecture.

\begin{figure}
\captionsetup[sub]{font=scriptsize}
    \centering
\begin{subfigure}
\centering
   \includegraphics[width=0.85\columnwidth]{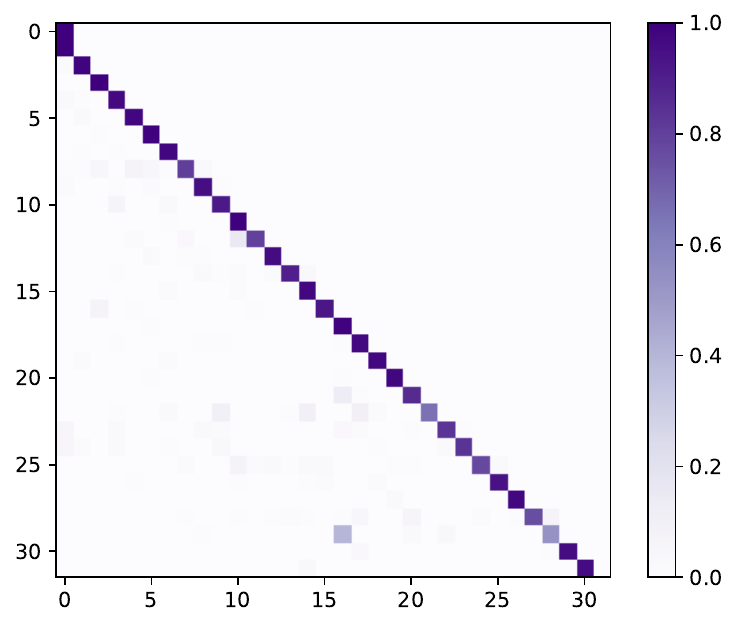} 
   \put(-114,-4){\fontsize{6}{3}\selectfont Index $i$}
      \put(-205,80){\rotatebox[origin=t]{90}{\fontsize{6}{3}\selectfont Index $n$}}
\end{subfigure}
\begin{subfigure}
\centering
   \includegraphics[width=0.85\columnwidth]{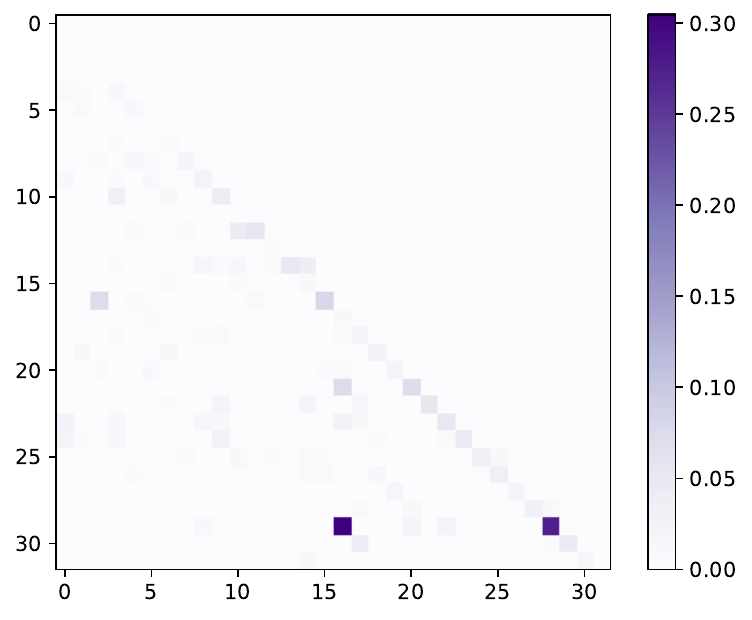} 
   \put(-117,-4){\fontsize{6}{3}\selectfont Index $i$}
      \put(-205,80){\rotatebox[origin=t]{90}{\fontsize{6}{3}\selectfont Index $n$}}
\end{subfigure}
\begin{subfigure}
\centering
   \includegraphics[width=0.85\columnwidth]{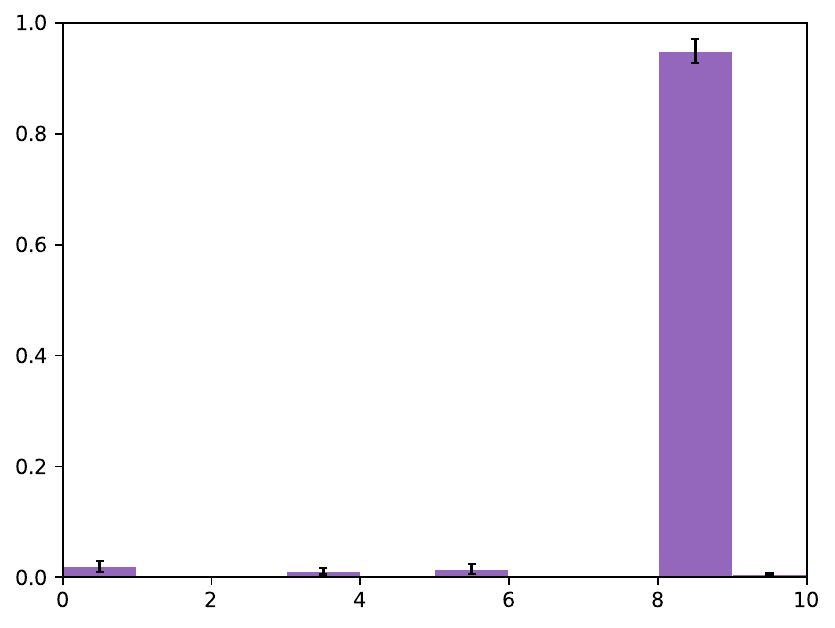} 
   \put(-115,-4){\fontsize{6}{3}\selectfont Sequence index $(i)$}
      \put(-205,77){\rotatebox[origin=t]{90}{\fontsize{5}{3}\selectfont Attention weight}}
\end{subfigure}
\caption{Attention matrix of the first attention layer, for a $2$-layer $1$-head transformer model trained on an order-$1$ Markov process, averaged across $100$ input sequences of length $128$. (a) and (b) plot the mean and standard deviation of the first 32 rows and columns of the attention matrix, while (c) zooms in on the column $n=10$ and plots the mean attention for this column. (a) and (c) show that for almost all indices $n$, the attention layer focuses only on the previous symbol $x_{n-1}$. (b) also shows that the attention pattern is largely independent of the actual sequence $\bm{x}$ considered, thereby providing evidence toward \Cref{ass:1}.}
\label{fig:4}
\end{figure}
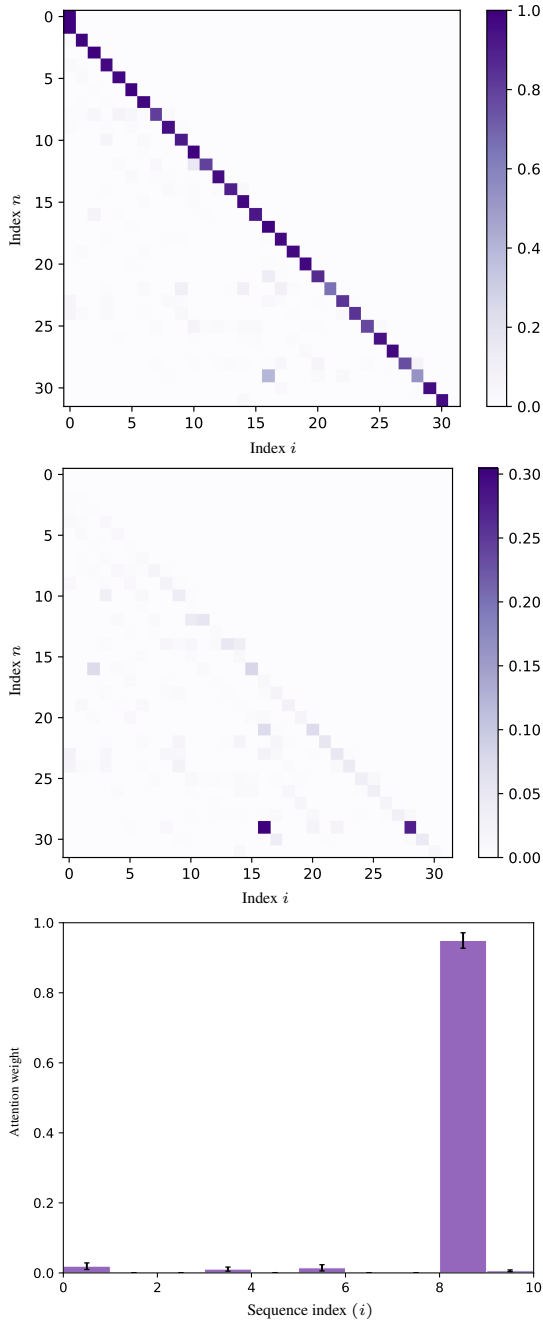

We prove conditional lower bounds under some natural assumptions on the nature of the attention patterns learnt by the transformer. To motivate these assumptions, consider the experiment in \Cref{fig:4}, where we train an attention-only transformer with $2$ layers and $1$ head, on data drawn from a random order-$1$ Markov process. At test-time, we sample sequences from the same process and plot the attention patterns learnt in the first layer of the model. Notice that the attention pattern learnt by the model at layer $1$ is largely independent of the input sequences themselves and only depends on the position. These, and additional experiments in \Cref{app:justification} support the assumption we make on the transformer in the size lower bound.

For higher values of $k$, we observe that there is some dependency of the learnt attention pattern on the inputs, corresponding to higher variance 
For higher values of $k$, we defer the experimental results to the appendix, since it requires plotting the attention across multiple heads.

\begin{assumption} \label{ass:1}
In an $L$-layer attention-only transformer with $H$ heads per layer, assume that layers $\ell = 1 ,2,\cdots,L-1$ and heads $h \in [H]$ realize an attention pattern where $\operatorname{att}_{n,i}^{(\ell,h)}$ only depends on the positions $n$ and $i$ and on $\ell$ and $h$, but not on the input sequence $x_1,\cdots,x_T$.
\end{assumption}

Rather than proving the size lower bound depending on the transformers ability to represent the conditional $k$-gram itself, we consider a simplification and assume that the goal of the model is to represent a $k^{\text{th}}$-order induction head (\Cref{def:k-order_IH}) in the last layer. Although learning a $k^{\text{th}}$-order induction head is not strictly necessary for the transformer to be able to represent conditional $k$-grams, note that every construction we have considered so far (cf. \Cref{theorem:simple,theorem:simple-k,theorem:logk,theorem:main}) go through this mechanism to realize the conditional $k$-gram model. Likewise, for other related problems, such as the causal learning task in \cite{nichani2024transformers}, the causal structure is captured by an extension of the $k^{\text{th}}$-order induction head to general causal graphs.

\begin{theorem} \label{theorem:lb-Llayer}
Consider an $L$-layer attention-only transformer with $1$ head per layer and relative position encodings, which satisfies \Cref{ass:1}. If $L \le 1 + \log_2 (k-2)$, the attention pattern in layer $L$ of the transformer cannot represent a $k^{\text{th}}$-order induction head.
\end{theorem}

While this lower bound is not unconditional, meaning that it does not directly imply that the transformer cannot represent conditional $k$-grams, it is important to understand the interpretation of this result: attention-only transformers which somehow break through this barrier need to use a significantly different mechanism to realize the conditional $k$-gram model. It is an interesting question to see if the size lower bound on representing \kth induction heads also necessarily implies a lower bound on representing the conditional $k$-gram model under the same assumptions. We also prove a size lower bound on $2$-layer transformer models under \Cref{ass:1}.

\begin{theorem} \label{theorem:lb-kheads}
Consider an $2$-layer attention-only transformer with $H$ heads in the first layer and relative position encodings, and assume that \Cref{ass:1} is satisfied. If $H \le k-3$, the attention pattern in the $2^{\text{nd}}$ layer cannot represent a $k^{\text{th}}$-order induction head.
\end{theorem}

This result implies that under \Cref{ass:1}, attention-only transformers indeed require the number of heads to scale linearly in $k$ to be able to represent a \kth induction head. In particular, under \Cref{ass:1} this implies that a $2$-layer attention-only transformer with $1$ head cannot realize a \kth induction head for any $k \ge 4$. Likewise, under the same assumption, a $3$-layer attention-only transformer with $1$ head cannot realize a \kth induction head for any $k \ge 6$. These results give more weight to the experiment in \Cref{fig:order8} where we observe that a $2$-layer transformer learns a \kth Markov process for $k = 4$ and a $3$-layer transformer learns a \kth Markov process for $k = 8$, and show that non-linearities in the architecture allow the transformer to break past the size barriers in \Cref{theorem:lb-Llayer,theorem:lb-kheads}.

Both of these theorems follow as a consequence of a more general size lower bound on attention-only transformers with $H_\ell$ heads in layer $\ell \in [L]$ which are able to represent a \kth induction head under \Cref{ass:1}. This general result also shows the utility in having deeper transformers, relative to wider (i.e., more number of heads) ones, in the context of learning \kth Markov processes.

\begin{theorem} \label{theorem:lb-general}
Consider an $L$-layer transformer with $h_{\ell}$ heads in layer $L$. Assuming the transformer satisfies \Cref{ass:1}, if $\prod_{\ell = 1}^{L-1} (H_{\ell} + 1) \le k-2$, the attention pattern in layer $L$ cannot represent a \kth induction head.
\end{theorem}

\section{Conclusion}
We observe empirically that $2$ and $3$ layer transformers are able to learn $k^{\text{th}}$-order Markov chains for much higher values of $k$ than previously anticipated. We show there are $O(\log(k))$-layer constructions of attention-only transformers which are able to learn the conditional $k$-gram model, which is the in-context MLE of the Markov model. With non-linearities in the model, we show that a $3$-layer $1$-head transformer is capable of representing the same. We show that $1$-layer transformers cannot represent conditional $k$-grams for any $k\ge3$ unless the number of heads or embedding dimension scale almost linearly in $T$. We also prove a conditional lower bound showing attention-only transformers need $\Omega (\log(k))$ layers to represent $k^{\text{th}}$-order induction heads, under an assumption on the realized attention patterns. Under the same assumptions, we show that a $2$-layer transformer needs $\Omega (k)$ heads in the first layer to realize a \kth induction head in the second layer. Our work focuses on the representational aspects of transformers; understanding the learning dynamics of gradient descent on these problems is an important next question.

%% file: appendix.tex
\newpage

\onecolumn
\appendix

\addcontentsline{toc}{section}{Appendix} 
\part{Appendix} 
\parttoc %

\paragraph{Notation.}
The notation $e_i^{d'} \in \mathbb{R}^{d'}$ refers to the one-hot encoding of $i$ in $d'$ dimensions. In other words it is the $i^{\text{th}}$ standard basis vector in $d'$ dimensions. The notation $\texttt{Blkdiag} (\{ A_1,A_1,\cdots,A_m \})$ refers to the block diagonal matrix with $i^{\text{th}}$ block as $A_i$.

\section{Proof of \Cref{theorem:simple}} \label{app:simple}

We will first prove \Cref{theorem:simple}. In the first layer, choose the embeddings as,
\begin{align}
    \bm{x}_n^{(1)} = \texttt{Emb} (x_n) = \kappa \begin{bmatrix} \bm{1}_{1 \times 2} & e_{x_n}^S & \bm{0}_{1 \times 2S} \end{bmatrix}^T \in \mathbb{R}^{d}.
\end{align}
for a constant $\kappa > 0$ to be chosen later and $d = 2S+2$. The relative position encodings will essentially be supported on the first two coordinates, the middle $S$ coordinates are a one-hot encoding of the symbol $x_n$ and the last $2S$ coordinates are $0$. The relative position encodings in the first layer are chosen to be $\bm{p}_{n-i}^{(1),K} = \kappa \left( -1 + \mathbb{I} (n-i=1 \}) \right) e_1^d \in \mathbb{R}^d$ and $\bm{p}_{n-i}^{(1),V} = \bm{0} \in \mathbb{R}^d$. Choose $\bm{W}_K^{(1)}$ and $\bm{W}_Q^{(1)}$ to be $e_1^d (e_1^d)^T \in \mathbb{R}^{d \times d}$. With this choice,
\begin{align}
    \left\langle \bm{W}_K^{(1)} \big( \bm{x}_i^{(1)} + \bm{p}_{n-i}^{(1),K} \big), \bm{W}_Q^{(1)} \bm{x}_n^{(1)} \right\rangle = \kappa \mathbb{I} (n-i=1)
\end{align}
\
As $\kappa \to \infty$, the attention pattern (which takes the softmax over of these inner products over $i \in [n]$) computes,
\begin{align}
    \operatorname{att}_{n, i}^{(1)} = \mathbb{I} (i = n-1)
\end{align}
for any $n > 1$. Choose the value matrix as,
\begin{align}
    \bm{W}_V^{(1)} = \begin{bmatrix} \bm{0}_{(2+S) \times 2} & \bm{0} & \bm{0} \\ \bm{0} & I_{S \times S} & \bm{0} \\ \bm{0} & \bm{0} & \bm{0} \end{bmatrix} \in \mathbb{R}^{d \times d}
\end{align}
And with this choice and the residual connection, we get,
\begin{align}
    \bm{x}_n^{(2)} = \kappa \begin{bmatrix} \bm{1}_{1 \times 2} & e^S_{x_n} & e^S_{x_{n-1}} & \bm{0} \end{bmatrix} \in \mathbb{R}^{d}
\end{align}
which serves as the input to the $2^{\text{nd}}$ transformer layer.

\paragraph{Layer 2.} In layer $2$, the relative position encodings $\bm{p}_{n-i}^{K,(2)}$ and $\bm{p}_{n-i}^{V,(2)}$ are all set as $0$. The key matrix picks out the $e_{x_n}^S$ block out of $\bm{x}_n^{(2)}$ and the query vector picks out the $e_{x_{i-1}}^S$ block out of $\bm{x}_{i-1}^{(2)}$. In particular, these matrices are chosen so that,
\begin{align}
\begin{split}
    \bm{W}_K^{(2)} \bm{x}^{(2)}_i &= \kappa \begin{bmatrix} \bm{1}_{1 \times 2} & e^S_{x_{i-1}} & \bm{0} \end{bmatrix}^T \in \mathbb{R}^d,\\
    \bm{W}_Q^{(2)} \bm{x}^{(2)}_n &= \kappa \begin{bmatrix} \bm{1}_{1 \times 2} & e^S_{x_n} & \bm{0} \end{bmatrix}^T \in \mathbb{R}^d
\end{split}
\end{align}
Taking the inner product of these vectors, and taking $\kappa \to \infty$, observe that the attention pattern concentrates on the uniform distribution over all coordinates $i$ such that $x_{i-1} = x_n$. More formally, the attention pattern for any $n > 1$ is,
\begin{align}
    \operatorname{att}_{n,i}^{(2)} =
    \frac{\mathbb{I} ( x_{i-1} = x_n )}{\sum_{i = 2}^n \mathbb{I} ( x_{i-1} = x_n )},
\end{align}
assuming $\sum_{i =2}^n \mathbb{I} ( x_{i-1} = x_n ) > 0$.
Having realized this attention pattern, may choose the value and subsequent linear layer appropriately. The value matrix simply picks out the $e_{x_i}^S$ block from $\bm{x}_i^{(2)}$ and places it into the last $S$ coordinates of $\bm{x}_i^{(3)}$, and the linear layer simply extracts this block and outputs it (after scaling down by a factor of $\kappa$), realizing the logits,
\begin{align}
    \operatorname{logit}_n = \frac{1}{\sum_{i=2}^n \mathbb{I} (x_{i-1} = x_n)} \sum_{i=2}^n \mathbb{I} (x_{i-1} = x_n) \cdot e_{x_i}^S.
\end{align}
if $\sum_{i=2}^n \mathbb{I} ( x_{i-1} = x_n) > 0$. In particular, under the same condition,
\begin{align}
    \operatorname{logit}_T (x_{T+1}) = 
    \frac{\sum_{n=2}^T \mathbb{I} (x_n = x_{T+1}, x_{n-1} = x_T)}{\sum_{i=2}^n \mathbb{I} (x_{n-1} = x_{T})}
\end{align}
assuming $\sum_{i=2}^n \mathbb{I} (x_{n-1} = x_{T})$, which is the conditional $1$-gram model.

\subsection{Extension to $k$-heads: Proof of \Cref{theorem:simple-k}} \label{app:simple-k}

In the first layer, the embeddings are chosen to be,
\begin{align}
    \bm{x}_n^{(1)} = \texttt{Emb} (x_n) = \kappa \left[ \begin{array}{c|c|c|c} \bm{0}_{1 \times k} & 1 & e_{x_n}^S & \bm{0}_{1 \times (k+1)S} \end{array} \right]^T \in \mathbb{R}^{d}
\end{align}
With $d = (k+1)(S+1)+S$. The relative position encodings are chosen as $\bm{p}_i^{K,(1)} = \begin{bmatrix} e_i^k & \bm{0} \end{bmatrix}^T$ for $1 \le i \le k$ and $\bm{p}_i^{K,(1)} = \bm{0}$ otherwise. Similarly, $\bm{p}_i^{V,(1)} = \bm{0}$ for every $i$. The $h^{\text{th}}$ head has key and query matrices,
\begin{align}
\begin{split}
    \bm{W}_Q^{(1,h)} &= \begin{bmatrix}
        \bm{0}_{1 \times k} & 1 & \bm{0} \\
        \bm{0} & \bm{0} & \bm{0}
    \end{bmatrix} \\
    \bm{W}_K^{(1,h)} &= \begin{bmatrix}
        \bm{0}_{1 \times (h-1)} & 1 & \bm{0} \\
        \bm{0} & \bm{0} & \bm{0}
    \end{bmatrix}
\end{split}
\end{align}
With these choices, and letting $\kappa \to \infty$, the $h^{\text{th}}$ layer computes the attention pattern,
\begin{align}
    \operatorname{att}_{n,i}^{(1,h)} = \mathbb{I} (i=n-h).
\end{align}
Choose the corresponding value matrix as,
\begin{align}
    \bm{W}_V^{(1,h)} = \begin{bmatrix}
        \bm{0}_{(2+hS) \times 2} & \bm{0} & \bm{0} \\
        \bm{0} & I_{S \times S} & \bm{0} \\
    \end{bmatrix}
\end{align}
choosing the projection matrix appropriately, the output of the transformer after the first residual connection is,
\begin{align} \label{eq:xn2}
    \bm{x}_n^{(2)} = \kappa \left[ \begin{array}{c|c|c|c|c}
        \bm{0}_{1 \times k} & 1 & e_{x_n}^S & \cdots & e_{x_{n-k}}^S
    \end{array} \right]^T.
\end{align}

\paragraph{Layer 2.} In this layer, the relative position encodings $\bm{p}_{n-i}^{K,(2)}$ and $\bm{p}_{n-i}^{V,(2)}$ are all set as $0$. The key and query matrices are chosen as,
\begin{align}
\begin{split}
\bm{W}_Q^{(2)} &= \begin{bmatrix}
    \bm{0}_{Sk \times k} & I_{(Sk+1) \times (Sk+1)} & \bm{0} \\
    \bm{0} & \bm{0} & \bm{0}
\end{bmatrix} \\
\bm{W}_K^{(2)} &= \begin{bmatrix}
    \bm{0}_{Sk \times (k+S)} & I_{(Sk+1) \times (Sk+1)} \\
    \bm{0} & \bm{0}
\end{bmatrix}.
\end{split}
\end{align}
With this choices, we have that,
\begin{align}
    \left\langle \bm{W}_K^{(2)} \bm{x}^{(2)}_i, \bm{W}_Q^{(2)} \bm{x}^{(2)}_n \right\rangle = \kappa \sum_{j=1}^k \mathbb{I} (x_{i-j} = x_{n-j+1}).
\end{align}
Taking $\kappa \to \infty$, observe that the attention pattern concentrates on the uniform distribution over all coordinates $i$ such that $x_{i-j} = x_{n-j+1}$ for all $j \in [k]$. More formally, if $\sum_{i=2}^n \mathbb{I} (x_{i-1} = x_n) > 0$, the attention pattern for any $n > 1$ is,
\begin{align}
    \operatorname{att}_{n,i}^{(2)} =
        \frac{\mathbb{I} (\forall j \in [k],\ x_{i-j} = x_{n-j+1})}{\sum_{i=k+1}^n \mathbb{I} (\forall j \in [k],\ x_{i-j} = x_{n-j+1})}.
\end{align}
The value matrix picks out $e_{x_i}^S$ from the embedding $\bm{x}_i^{(2)}$ (\cref{eq:xn2}) and places it in the last $S$ coordinates. The subsequent linear layer picks out the last $S$ coordinates, resulting in the logits,
\begin{align}
    \operatorname{logit}_n = \sum_{i=k+1}^n \frac{\mathbb{I} (\forall j \in [k],\ x_{i-j} = x_{n-j+1})}{\sum_{i=k+1}^n \mathbb{I} (\forall j \in [k],\ x_{i-j} = x_{n-j+1})} e_{x_i}^S,
\end{align}
assuming that $\sum_{i=k+1}^n \mathbb{I} (\forall j \in [k],\ x_{i-j} = x_{n-j+1}) > 0$.
In particular,
\begin{align}
    \operatorname{logit}_T (x_{T+1}) =
    \frac{\sum_{n=k+1}^T \mathbb{I} (\forall 0 \le j \le k,\ x_{n-j} = x_{T-j+1})}{\sum_{n=k+1}^T \mathbb{I} (\forall 1 \le j \le k,\ x_{n-j} = x_{T-j+1})},
\end{align}
assuming $\sum_{n=k+1}^T \mathbb{I} (\forall 1 \le j \le k,\ x_{n-j} = x_{T-j+1}) > 0$, i.e., the conditional $k$-gram model.

\section{Proof of \Cref{theorem:logk}} \label{app:logk}

Define $k^\star = 2^{\lceil \log_2 (k+1) \rceil}$ by rounding $k+1$ up to the nearest power of $2$ and $\ell^\star = \log_2 (k^\star)$. In the setting of relative position encodings, given the sequence $x_1,\cdots,x_n$, while generating the output of the attention + feedforward layer for the symbol $x_n$, the embeddings $\bm{x}_i = \texttt{Emb} (x_n) + \bm{p}_{n-i}$ are used for $i \in [n]$. In other words, the position encoding vector is taken relative to the end of the sequence, rather than the start of the sequence. Consider the embedding of $x$ as,
\begin{align}
    \bm{x}_n^{(1)} = \texttt{Emb} (x_n) = \left[ \begin{array}{c|c|c|c|c}
        \bm{0}_{1 \times \ell^\star} & 1 & e_{x_n}^S & \bm{0}_{1 \times (k^\star-1) S} & \bm{0}_{1 \times S}
    \end{array} \right]^T \in \mathbb{R}^{(k^\star+1) S + \ell^\star+1}
\end{align}
where $e_i^{d'} \in \mathbb{R}^S$ is the standard basis vector in $d'$ dimensions. And the relative position encoding for the keys as,
\begin{align}
    \bm{p}_i^{(1),K} = \begin{cases} \begin{bmatrix} \bm{1}_{1 \times \ell^\star} & \bm{0} \end{bmatrix}^T, \qquad &\text{if } i=0, \\
    \begin{bmatrix} e^{\ell^\star}_{1 + \log_2 (i)} & \bm{0} \end{bmatrix}^T &\text{if } i \in \{ 1,2,4,\cdots,k^\star/2 \} \\
    \ \bm{0}_{d \times 1} &\text{otherwise.}
    \end{cases} \label{eq:pi}
\end{align}
And for the value vectors, $\bm{p}_i^V = \bm{0}$ for all $i$.

For the first layer and first head, we will describe the value, key and query matrices. Choose,
\begin{align}
\begin{split}
    \bm{W}^{(1)}_K &= \sqrt{\kappa} \begin{bmatrix} 1 & \bm{0} \\ \bm{0} & \bm{0} \end{bmatrix}, \text{ and,} \\
    \bm{W}_Q^{(1)} &= \sqrt{\kappa} \begin{bmatrix}
        \bm{0}_{1 \times l^\star} & 1 & \bm{0} \\
        \bm{0} & \bm{0} & \bm{0}
    \end{bmatrix}.
\end{split} \label{eq:WKQ1}
\end{align}
Then, observe that for $i \ge 1$,
\begin{align} \nonumber
    \left\langle \bm{W}_K^{(1)} \big( \bm{x}_{n-i} + \bm{p}_i^{(1),K} \big), \bm{W}_Q^{(1)} \bm{x}_{n} \right\rangle = \kappa \mathbb{I}(i=1)
\end{align}
and for $i = 0$,
\begin{align} \nonumber
    \left\langle \bm{W}_K^{(1)} \big( \bm{x}_n + \bm{p}_0^{(1),K} \big), \bm{W}_Q^{(1)} \bm{x}_{n} \right\rangle = \kappa
\end{align}
In particular, letting $\kappa \to \infty$, the attention pattern is,
\begin{align} 
    \operatorname{att}^{(1)}_{n,n-i} = \frac{1}{2} \mathbb{I} (i = 0) + \frac{1}{2} \mathbb{I} (i = 1).
\end{align}

Choose the value matrix as,
\begin{align} \nonumber
    \bm{W}_V^{(1)} = \begin{bmatrix}
        \bm{0}_{(\ell^\star+S) \times \ell^\star} & \bm{0} \\ \bm{0} & 2 I
    \end{bmatrix}
\end{align}
together with the residual connection, we get,
\begin{align}
    \bm{x}_n^{(2)} = \left[ \begin{array}{c|c|c|c|c|c}
        \bm{0}_{1 \times \ell^\star} & 1 & e_{x_n}^S & e_{x_n}^S + e_{x_{n-1}}^S & \bm{0}_{1 \times (k^\star-2) S} & \bm{0}_{1 \times S}
    \end{array} \right]^T
\end{align}

\paragraph{Layer $\ell+1$.} By induction, assume that the output of the $\ell^{\text{th}}$ transformer layer is of the form,
\begin{align}
    \bm{x}_n^{(\ell+1)} = \left[ \begin{array}{c|c|c|c|c}
        \bm{0}_{1 \times \ell^\star} & 1 & \bm{v}_n &
        \bm{0}_{1 \times (k^\star-2^\ell)S} & \bm{0}_{1 \times S}
    \end{array} \right]^T
\end{align}
for some vector $\bm{v}_n \in \mathbb{R}^{2^\ell S}$. We will show that with appropriately chosen key, query and value vectors in the $(\ell+1)^{\text{th}}$ layer, the output of this layer is,
\begin{align}
    \bm{x}_n^{(\ell+2)} = \left[ \begin{array}{c|c|c|c|c|c}
        \bm{0}_{1 \times \ell^\star} & 1 & \bm{v}_n  & \bm{v}_n + \bm{v}_{n - 2^\ell} &
        \bm{0}_{1 \times (k^\star-2^{\ell+1})S} & \bm{0}_{1 \times S}
    \end{array} \right]^T
\end{align}
We will consider the same relative position encodings and query matrix in this layer as in the first layer (\cref{eq:pi,eq:WKQ1}). Consider a key matrix of the form,
\begin{align} \nonumber
        \bm{W}_K^{(\ell+1)} = \begin{bmatrix}
            \bm{0}_{1 \times \ell} & \sqrt{\kappa} & \bm{0} \\
            \bm{0} & \bm{0} & \bm{0}
        \end{bmatrix}
\end{align}
With this choice, observe that for $i \ge 1$,
\begin{align}
    \left\langle \bm{W}_K^{(\ell+1)} \big( \bm{x}_{n-i}^{(\ell+1)} + \bm{p}_i^{(\ell+1),K} \big), \bm{W}_Q^{(\ell+1)} \bm{x}_{n}^{(\ell+1)} \right\rangle = \kappa \cdot \mathbb{I} (i = 2^\ell) \nonumber
\end{align}
and for $i = 0$,
\begin{align} \nonumber
    \left\langle \bm{W}_K^{(\ell+1)} \big(\bm{x}_{n}^{(\ell+1)} + \bm{p}_0^{(\ell+1),K} \big), \bm{W}_Q^{(\ell+1)} \bm{x}_{n}^{(\ell+1)} \right\rangle = \kappa
\end{align}

In particular, letting $\kappa \to \infty$, the attention pattern is,
\begin{align}
    \operatorname{att}_{n,n-i}^{(\ell+1)} = \frac{1}{2} \mathbb{I} (i = 0) + \frac{1}{2} \mathbb{I} (i= 2^\ell).
\end{align}
Choosing the value matrix as,
\begin{align*}
    \bm{W}_V^{(\ell+1)} = \begin{bmatrix}
        \bm{0}_{(\ell^\star + 2^\ell S) \times \ell^\star} & \bm{0} \\ \bm{0} & 2 I
    \end{bmatrix}, 
\end{align*}
we get,
\begin{align}
    \bm{x}_n^{(\ell+2)} = \left[ \begin{array}{c|c|c|c|c|c}
        \bm{0}_{1 \times \ell^\star} & 1 & \bm{v}_n & \bm{v}_n + \bm{v}_{n-2^\ell} & \bm{0}_{1 \times (k^\star-2^{\ell+1}) S} & \bm{0}_{1 \times S}
    \end{array} \right]^T \label{eq:88}
\end{align}

\paragraph{Final last transformer layer ($\ell = \ell^\star$).} The output of the second last transformer layer, indexed $\ell^\star-1$ is,
\begin{align}
    \bm{z}_n^{(\ell^\star)} \triangleq \bm{x}_n^{(\ell^\star)} &= \left[ \begin{array}{c|c|c|c|c}
        \bm{0}_{1 \times \ell^\star} & 1 & \bm{v}_n^{(\ell^\star-1)} &
        \bm{v}_n^{(\ell^\star-1)} + \bm{v}_{n-2^{\ell^\star-1}}^{(\ell^\star-1)} & \bm{0}_{1 \times S}
    \end{array} \right]^T \nonumber\\
    &= \left[ \begin{array}{c|c|c|c|c}
        \bm{0}_{1 \times \ell^\star} & 1 & \bm{v}_n^{(\ell^\star-1)} &
        \bm{v}_n^{(\ell^\star-1)} + \bm{v}_{n-\frac{k^\star}{2}}^{(\ell^\star-1)} & \bm{0}_{1 \times S}
    \end{array} \right]^T,\nonumber
\end{align}
which follows by plugging in the definition of $k^\star$. Note that there exists a linear transformation $\bm{L}^{(\ell^\star)}$ such that,
\begin{align}
    \bm{z}_n^{(\ell^\star-1)} \triangleq \bm{L}^{(\ell^\star)} \bm{x}_n^{(\ell^\star)} = \left[ \begin{array}{c|c|c|c|c}
        \bm{0}_{1 \times \ell^\star} & 1 & \bm{v}_n^{(\ell^\star-1)} & \bm{v}_{n-\frac{k^\star}{2}}^{(\ell^\star-1)} & \bm{0}_{1 \times S}
    \end{array} \right]^T \nonumber
\end{align}
This can be further decomposed as,
\begin{align}
    &\bm{z}_n^{(\ell^\star-1)} \nonumber\\
    &= \left[ \begin{array}{c|c|c|c|c|c|c}
        \bm{0}_{1 \times \ell^\star} & 1 &
        \bm{v}_{n}^{(\ell^\star-2)} &
        \bm{v}_{n}^{(\ell^\star-2)} + \bm{v}_{n-2^{\ell^\star-2}}^{(\ell^\star-2)} &
        \bm{v}_{n - \frac{k^\star}{2}}^{(\ell^\star-2)} &
        \bm{v}_{n - \frac{k^\star}{2}}^{(\ell^\star-2)} + \bm{v}_{n-\frac{k^\star}{2} - 2^{\ell^\star-2}}^{(\ell^\star-2)} & \bm{0}_{1 \times S}
    \end{array} \right]^T \nonumber
\end{align}
And yet again there exists a linear transformation $\bm{L}^{(\ell^\star-1)}$ which transforms this as,
\begin{align}
    &\bm{z}_n^{(\ell^\star-2)} \triangleq \bm{L}^{(\ell^\star-1)} \bm{z}_n^{(\ell^\star-1)} \nonumber\\
    &\quad= \left[ \begin{array}{c|c|c|c|c|c|c}
        \bm{0}_{1 \times \ell^\star} & 1 &
        \bm{v}_{n}^{(\ell^\star-2)} &
        \bm{v}_{n-2^{\ell^\star-2}}^{(\ell^\star-2)} &
        \bm{v}_{n - \frac{k^\star}{2} - 2^{\ell^\star-2}}^{(\ell^\star-2)} &
        \bm{v}_{n - \frac{k^\star}{2} - 2^{\ell^\star-2}}^{(\ell^\star-2)} & \bm{0}_{1 \times S}
    \end{array} \right]^T \nonumber\\
    &\quad= \left[ \begin{array}{c|c|c|c|c|c|c}
        \bm{0}_{1 \times \ell^\star} & 1 &
        \bm{v}_{n}^{(\ell^\star-2)} &
        \bm{v}_{n-\frac{k^\star}{4}}^{(\ell^\star-2)} &
        \bm{v}_{n - \frac{k^\star}{2}}^{(\ell^\star-2)} &
        \bm{v}_{n-\frac{3k^\star}{4}}^{(\ell^\star-2)} & \bm{0}_{1 \times S}
    \end{array} \right]^T
\end{align}
By recursing this argument and composing all the linear transformations, up to a global permutation, we get that,
\begin{align}
    \prod_{\ell=1}^{\ell^\star} \bm{L}^{(\ell)} \bm{x}_n^{(\ell^\star)} &= \left[ \begin{array}{c|c|c|c|c|c|c}
        \bm{0}_{1 \times \ell^\star} & 1 & \bm{v}_n^{(1)} & \bm{v}_{n-1}^{(1)} & \cdots & \bm{v}_{n-(k^\star-1)}^{(1)} & \bm{0}_{1 \times S}
    \end{array} \right]^T \nonumber\\
    &= \left[ \begin{array}{c|c|c|c|c|c}
        \bm{0}_{1 \times \ell^\star} & 1 & e_{x_n}^S & \cdots & e_{x_{n-(k^\star-1)}}^S & \bm{0}_{1 \times S}
    \end{array} \right]^T \label{eq:33}
\end{align}
In the final layer, we will right multiply the key, query and value matrices by $\bm{L}^\star = \prod_{\ell=1}^{\ell^\star} \bm{L}^{(\ell)}$. The effect can be interpreted as operating the original key, query and value matrices on the embedding vectors in \cref{eq:33}. In the final layer, we will set all the position encodings to be $\bm{0}$ and consider the key and query matrices,
\begin{align}
    \begin{split}\bm{W}_K^{(\ell^\star)} &= \sqrt{\kappa} \begin{bmatrix}
        \bm{0}_{Sk \times (\ell^\star+1+S)} & I_{Sk \times S k} & \bm{0} \\
        \bm{0} & \bm{0} & \bm{0}
    \end{bmatrix} \\
    \bm{W}_Q^{(\ell^\star)} &= \sqrt{\kappa} \begin{bmatrix}
        \bm{0}_{Sk \times (\ell^\star+1)} & I_{Sk \times S k} & \bm{0} \\
        \bm{0} & \bm{0} & \bm{0}
    \end{bmatrix}
    \end{split}
\end{align}
Then,
\begin{align}
    \left\langle \bm{W}_K^{(\ell^\star)} \bm{L}^\star \bm{x}_{n-i}^{(\ell^\star)}, \bm{W}_Q^{(\ell^\star)} \bm{L}^\star \bm{x}_n^{(\ell^\star)} \right\rangle &= \kappa \sum_{j=0}^{k-1} \mathbb{I} ( x_{n-j} = x_{i-1-j} )
\end{align}
Where we must be careful to note that the input $\bm{x}_n^{(\ell^\star)}$ contains copies of $e_{x_n}, e_{x_{n-1}},\cdots,e_{x_{n-k}}$ since $k^\star \ge k+1$ by definition.

Letting $\kappa \to \infty$, if there exists $i$ such that $\sum_{j=0}^{k-1} \mathbb{I} (x_{n-j} = x_{i-j-1}) > 0$, for $n \ge k$, the attention pattern is,
\begin{align}
    \operatorname{att}_{n,i}^{(\ell^\star)} =
    \frac{\mathbb{I} ( x_{i-1} = x_n , x_{i-2} = x_{n-1}, \cdots, x_{i-k} = x_{n-k+1})}{\sum_{i = k}^n \mathbb{I} ( x_{i-1} = x_n , x_{i-2} = x_{n-1}, \cdots, x_{i-k} = x_{n-k+1}) }
\end{align}
Finally, choose,
\begin{align}
    \bm{W}_V^{(\ell^\star+2)} = \begin{bmatrix}
        \bm{0}_{(d-S) \times (\ell^\star+1)} & \bm{0} & \bm{0} \\
        \bm{0}_{S \times (\ell^\star+1)} & I_{S \times S} & \bm{0}
    \end{bmatrix},
\end{align}
we get,
\begin{align}
    \bm{x}_n^{(\ell^\star+1)} + \sum_{i=k}^n \frac{\mathbb{I} ( x_{i-1} = x_n , x_{i-2} = x_{n-1}, \cdots, x_{i-k} = x_{n-k+1})}{\sum_{i = k}^n \mathbb{I} ( x_{i-1} = x_n , x_{i-2} = x_{n-1}, \cdots, x_{i-k} = x_{n-k+1}) } \begin{bmatrix} \bm{0}_{(d-S) \times 1} \\ e_{x_i} \end{bmatrix}
\end{align}
Choosing the subsequent linear layer as,
\begin{align}
    \bm{A} &= \begin{bmatrix}
        \bm{0}_{S \times (d-S)} & I_{S \times S}
    \end{bmatrix} \\
    \bm{b} &= \bm{0}_{S \times 1}
\end{align}
Results in the output,
\begin{align}
    \operatorname{logit}_T (x_{T+1}) = \sum_{n=k}^T \frac{\mathbb{I} ( x_n = x_{T+1}, x_{n-1} = x_T , x_{n-2} = x_{T-1}, \cdots, x_{n-k} = x_{T-k+1})}{\sum_{n = k}^T \mathbb{I} ( x_{n-1} = x_T , x_{n-2} = x_{T-1}, \cdots, x_{n-k} = x_{T-k+1}) }
\end{align}
which is precisely the in-context conditional $k$-gram.

\section{Proof of \Cref{theorem:main}} \label{app:constant}

\subsection{Modifying the definition of layer normalization} \label{sec:trick}

In every layer, we will perform a simple transformation which is to double the hidden dimension $d$ and add a copy of $-\bm{x}_n^{(\ell)}$ into the last $d$ coordinates. This is possible by modifying the weights of the transformer appropriately as discussed below. A consequence of this transformation is that the feature mean of the $\bm{x}_n$'s is $\mu_n = 0$, and therefore the standard deviation $\sigma_n$ simply normalizes by the $L_2$-norm of the features. In order to avoid having to explicitly state this transformation at each layer, we will simply redefine the layer norm LN to output $\bm{v} / \| \bm{v} \|_2$ for the input vector $\bm{v}$, which is realized on the first $d$ coordinates of the transformed embeddings.

This transformation can be realized automatically by redefining the initial embeddings $\texttt{Emb} (x_n)$, and modifying the weights of the attention and feedforward subnetworks as follows: The input embeddings are changed to $\begin{bmatrix}
\texttt{Emb} (x_n) & - \texttt{Emb} (x_n)
\end{bmatrix}^T \in \mathbb{R}^{2d}$. The key and query matrices are chosen to be $0$ on the last $d$ coordinates in every layer; the value matrix for $i \ge 1$ is transformed to $\texttt{Blkdiag} \big( \big\{ \bm{W}_V^{(\ell)} , \bm{W}_V^{(\ell)} \big\} \big)$, and likewise changing the feedforward layer to the block diagonal matrices $\texttt{Blkdiag} \big( \big\{ \bm{W}_1^{(\ell)} , \bm{W}_1^{(\ell)} \big\} \big)$ and $\texttt{Blkdiag} \big( \big\{ \bm{W}_2^{(\ell)} , \bm{W}_2^{(\ell)} \big\} \big)$. This transformation adds a copy of $-\bm{x}_n^{(\ell)}$ into the last $d$ coordinates of the corresponding embeddings. 

\subsection{Notation and supplementary lemmas}

For each $i \in [T]$, define,
\begin{align}
    \bm{v}_i &= e_{x_{i-1}} + 3 \cdot e_{x_{i-2}} + \cdots + 3^{k-1} \cdot e_{x_{i-k}} \\
    \bm{u}_i &= e_{x_{i}} + 3 \cdot e_{x_{i-1}} + \cdots + 3^{k-1} \cdot e_{x_{i-k+1}}
\end{align}
Note that although $\bm{v}_i = \bm{u}_{i-1}$, we make the distinction between the two to avoid any confusion in what is stored in the embedding vector at time $i$ and at time $i-1$. Furthermore, define,
\begin{align}
    \mathcal{I}_n = \{ k+1 \le i \le n : \forall j \in [k], x_{i-j} = x_{n-j+1} \}.
\end{align}

\begin{lemma} \label{lemma:normerror}
If $i \in \mathcal{I}_n$, $\bm{z}_i = \bm{z}_{n-1}$. However, if $i \ge k+1$ but $i \not\in \mathcal{I}_n$, then, $\big\| \frac{\bm{v}_i}{\| \bm{v}_i \|_2} - \frac{\bm{u}_n}{\| \bm{u}_n \|_2} \big\|_2 \ge 3^{-k}$.
\end{lemma}

Let $j^\star \in \{ 0,1,\cdots,k-1\}$ denote the largest index $j$ such that $x_{n-j} \ne x_{i-j-1}$. Consider the coordinates $a = x_{n-j^\star} \in [S]$ and $b = x_{i-j^\star-1} \in [S]$. Then,
\begin{align}
    \langle \bm{v}_n, e_{a} \rangle - \langle \bm{u}_i, e_{a} \rangle &\ge 3^j - \sum_{j = 0}^{j^\star-1} 3^{j} = \frac{3^{j^\star}}{2}, \\ \langle \bm{u}_i, e_{b} \rangle - \langle \bm{v}_n, e_{b} \rangle &\ge \frac{3^{j^\star}}{2}
\end{align}
If $\| \bm{v}_n \|_2 \ge \| \bm{u}_i \|_2$, then,
\begin{align}
    \left\langle \frac{\bm{u}_i}{\| \bm{u}_i \|_2}, e_{b} \right\rangle - \left\langle \frac{\bm{v}_n}{\| \bm{v}_n\|_2}, e_{b} \right\rangle \ge \frac{\langle \bm{u}_i, e_{b} \rangle - \langle \bm{v}_n, e_{b} \rangle}{\max \{ \| \bm{u}_i \|_2, \| \bm{v}_n \|_2 \}} \ge \frac{3^{j^\star}}{2 \cdot \frac{3^k}{2}} = 3^{j^\star-k}
\end{align}
This uses the fact that $\bm{u}_i$ and $\bm{v}_n$ are coordinate-wise non-negative.
On the other hand, if $\| \bm{v}_n \|_2 \le \| \bm{u}_i \|_2$, using a similar analysis,
\begin{align}
    \left\langle \frac{\bm{u}_i}{\| \bm{u}_i \|_2}, e_{a} \right\rangle - \left\langle \frac{\bm{v}_n}{\| \bm{v}_n\|_2}, e_{a} \right\rangle \ge 3^{j^\star-k}.
\end{align}
In either case, there is a coordinate ($a$ or $b$) such that, $\bm{u}_i / \| \bm{u}_i \|_2$ and $\bm{v}_n / \| \bm{v}_n \|_2$ differ by at least $3^{j^\star-k}$. This implies the lower bound on the $L_2$ norm of the difference of the vectors.


\subsection{Proof of \Cref{theorem:main}}

Choose the input embeddings as, 
\begin{align}
    \bm{x}^{(1)}_n = \texttt{Emb} (x_n) = \begin{bmatrix}
        \bm{0}_{1 \times 3} & e_{x}^S & \bm{0}_{1 \times 5S}
    \end{bmatrix}^T \in \mathbb{R}^{6S + 3}
\end{align}
In the first two layers we will use the same relative position embeddings, in particular,
\begin{align}
    \bm{p}_{i}^{(1),K} = \bm{p}_{i}^{(2),K} = \begin{cases}
        \sqrt{\log (3)} \cdot \begin{bmatrix} 1 & \bm{0} \end{bmatrix}^T, &\text{if } i=0,\\
    (i+1) \sqrt{\log (3)} \cdot \begin{bmatrix} 0 & 1 & \bm{0} \end{bmatrix}^T, & \text{if } i \in \{ 1,2,\cdots,k-1\}, \\
    (k+1) \sqrt{\log (3)} \cdot \begin{bmatrix} 0 & 0 & 1 & \bm{0} \end{bmatrix}^T, \quad &\text{if } i=k.
    \end{cases},
    \label{eq:pemb}
\end{align}
and the value embeddings,
\begin{align}
    \bm{p}_i^{(1),V} = \bm{p}_i^{(2),V} = \begin{cases}
        3^i \begin{bmatrix}
    1 & \bm{0}
\end{bmatrix}^T \quad &\text{for } i \le k \\
    \bm{0} & i > k.
    \end{cases}
\end{align}
In the final layer, we will drop all position-related information and choose $\bm{p}_i^{(3),K} = \bm{p}_i^{(3),V} = \bm{0}$ for all $i$.

\paragraph{Layer 1.}  Consider the key and query matrices,
\begin{align}
\begin{split}
    \bm{W}^{(1)}_K &= \sqrt{\kappa} \cdot \begin{bmatrix} \bm{1}_{1 \times 2} & \bm{0} \\ \bm{0} & \bm{0} \end{bmatrix} \\
    \bm{W}_Q^{(1)} &= \sqrt{\kappa} \cdot \begin{bmatrix} \bm{0}_{1 \times 3} & \bm{1}_{1 \times S} & \bm{0} \\ \bm{0} & \bm{0} & \bm{0} \end{bmatrix}
\end{split}
\end{align}
Then, observe that,
\begin{align} \nonumber
    \left\langle \bm{W}_K^{(1)} \big( \texttt{Emb} (x_{n-i}) + \bm{p}_i^{(1),K} \big), \bm{W}_Q^{(1)} \texttt{Emb} (x_n) \right\rangle = \kappa (i+1) \log (3) \cdot \mathbb{I}(0 \le i \le \min \{ n, k \}-1)
\end{align}
Letting $\kappa \to \infty$, this results in the attention pattern,
\begin{align} \label{eq:l1att}
    \operatorname{att}^{(1)}_{n,n-i} = \frac{3^i \mathbb{I} (0 \le i \le \min \{n,k\}-1)}{\sum_{i'=0}^{\min \{ n, k \}-1} 3^{i'}}
\end{align}
Choose the value matrix as,
\begin{align} \nonumber
    \bm{W}_V^{(1)} = \begin{bmatrix}
        \bm{0}_{(S+3) \times 3} & \bm{0} \\
        \bm{0} & I
    \end{bmatrix}
\end{align}
The output of the attention layer (with the residual connection) is,
\begin{align}
    \widetilde{\bm{x}}_n^{(1)} &= \left[ \begin{array}{c|c|c|c}
        \bm{0}_{1 \times 3} & e_{x_n}^S & \bm{u}_n & \bm{0}_{1 \times 3S}
    \end{array} \right]^T, \text{ where, } \bm{u}_n = \sum_{i=0}^{\min \{n,k\}-1} \operatorname{att}_{n,n-i} e^S_{x_{n-i}}.
\end{align}
In the feedforward layer to follow, we will choose,
\begin{align}
\begin{split}
    \bm{W}_1^{(1)} &= I \\
    \bm{W}_2^{(1)} &= \begin{bmatrix}
    \bm{0}_{(3+2S) \times (3+S)} & \bm{0} & \bm{0} \\ \bm{0} & I_{S \times S} & \bm{0} \\
    \bm{0} & \bm{0} & \bm{0}
\end{bmatrix}
\end{split}
\end{align}
Which simply extracts $\bm{u}_n$ from $\widetilde{\bm{x}}_n^{(1)}$. With the subsequent layer norm and residual connection, the output of the first layer is,
\begin{align}
    \bm{x}_n^{(2)} = \left[ \begin{array}{c|c|c|c|c}
        \bm{0}_{1 \times 3} & e_{x_n}^S & \bm{u}_n & \frac{\bm{u}_n}{\| \bm{u}_n \|_2} & \bm{0}_{1 \times 3S}
    \end{array} \right]^T
\end{align}

\paragraph{Layer 2.}
In this layer, the relative position encodings and query matrix are the same as in layer $1$ but the key matrix is chosen as,
\begin{align}
    \bm{W}_K^{(2)} = \sqrt{\kappa} \begin{bmatrix}
        0 & \bm{1}_{1 \times 2} & \bm{0} \\
        \bm{0} & \bm{0} & \bm{0}
    \end{bmatrix}
\end{align}
With this choice, observe that,
\begin{align}
    \left\langle \bm{W}_K^{(2)} (\bm{x}^{(2)}_{n-i} + \bm{p}_i^{(1),K}), \bm{W}_Q^{(2)} \bm{x}_n^{(2)} \right\rangle = \kappa (i+1) \log (3) \cdot \mathbb{I}(1 \le i \le k)
\end{align}
As before, since $\kappa \to \infty$, this results in the attention pattern,
\begin{align}
    \operatorname{att}^{(2)}_{n,n-i} = \frac{3^i \mathbb{I} (1 \le i \le \min \{ k, n-1\})}{\sum_{i'=1}^{\min \{k, n-1\}} 3^{i'}}
\end{align}
which is similar, but subtly different from the attention pattern in the first layer (\cref{eq:l1att}). The first layer focuses on indices $n-i$ such that $0 \le i \le k-1$, while this layer focuses on $1 \le i \le k$. Choosing the value and projection matrices as,
\begin{align}
    \bm{W}_V^{(2)} = \begin{bmatrix}
        I_{3 \times 3} & \bm{0} & \bm{0} \\
        \bm{0}_{3S \times 3} & \bm{0} & \bm{0} \\
        \bm{0} & I_{S \times S} & \bm{0} \\
        \bm{0} & \bm{0} & \bm{0} \\
    \end{bmatrix}
\end{align}
The output of the attention layer (with the first residual connection) is,
\begin{align}
\begin{split}
    \widetilde{\bm{x}}_n^{(2)} &= \left[ \begin{array}{c|c|c|c|c|c|c}
        Z_n & \bm{0}_{1 \times 2} & e_{x_n}^S & \bm{u}_n & \frac{\bm{u}_n}{\| \bm{u}_n \|_2} & \bm{v}_n & \bm{0}_{1 \times 2S}
    \end{array} \right]^T, \\
    &\text{where, } \bm{v}_n = \sum_{i=1}^{\min \{k,n-1\}} \operatorname{att}_{n,n-i} e^S_{x_{n-i}},\\
    &\text{and, } Z_n = \sum_{i=1}^{\min \{k,n-1\}} \operatorname{att}_{n,n-i} 3^i,
    \end{split}
\end{align}
It is a short calculation to see that $Z_n = 3^{k+1}/5$ if $n \ge k+1$ and otherwise, $Z_n \le 3^k/5$. This will be useful later, since the value of $Z_n$ can be used to determine whether $n \ge k+1$ or $n \le k$ which will allow the the next layer to avoid calculating the attention at $i \le k$, where the evaluation $x_n = x_{i-1}, \cdots, x_{n-k+1} = x_{i-k}$ is not well defined. In the subsequent FFN layer, we will choose,
\begin{align}
\begin{split}
    \bm{W}_1^{(2)} &= I \\
    \bm{W}_2^{(2)} &= \begin{bmatrix}
    \bm{0}_{(3+4S) \times (3+3S)} & \bm{0} & \bm{0} \\ \bm{0} & I_{S \times S} & \bm{0} \\
    \bm{0} & \bm{0} & \bm{0}_{S \times 2S}
\end{bmatrix}
\end{split}
\end{align}
Which extracts $\bm{v}_n$ from the embedding $\widetilde{\bm{x}}_n^{(2)}$. 
With the layer norm and adding the final residual connection, the output of this layer is,
\begin{align}
    \bm{x}_{n}^{(3)} = \left[ \begin{array}{c|c|c|c|c|c|c|c}
        Z_n & \bm{0}_{2 \times 1} &
        e_{x_n}^S &
        \bm{u}_n &
        \frac{\bm{u}_n}{\| \bm{u}_n \|_2} &
        \bm{v}_n & 
        \frac{\bm{v}_n}{\| \bm{v}_n \|_2} & \bm{0}_{S \times 1}
    \end{array} \right]^T
\end{align}

\paragraph{Layer 3.} In this layer, all the relative position encodings are set as $\bm{0}$ and instead,
\begin{align}
    \begin{split}
    \bm{W}_Q^{(3)} &= \sqrt{2\kappa} \begin{bmatrix}
        1 & \bm{0} & \bm{0} & \bm{0} \\
        \bm{0} & \bm{0}_{S \times (2+3S)} & I_{S \times S} & \bm{0} \\
        \bm{0} & \bm{0} & \bm{0} & \bm{0}
    \end{bmatrix} \\
    \bm{W}_K^{(3)} &= \sqrt{2\kappa} \cdot \begin{bmatrix}
        1 & \bm{0} & \bm{0} & \bm{0} \\
        \bm{0} & \bm{0}_{S \times (2+4S)} & I_{S \times S} & \bm{0} \\
        \bm{0} & \bm{0} & \bm{0} & \bm{0}
    \end{bmatrix}
    \end{split}
\end{align}
With these choices,
\begin{align}
    \left\langle \bm{W}_K^{(3)} \bm{x}_i^{(3)} , \bm{W}_Q^{(3)} \bm{x}_n^{(3)} \right\rangle &= 2 \kappa Z_i Z_n + \frac{2 \kappa \langle \bm{v}_i, \bm{u}_n \rangle}{\| \bm{v}_i \|_2 \cdot \| \bm{u}_n \|_2} \nonumber \\
    &= 2 \kappa Z_i Z_n + 2 \kappa - \kappa \left\| \frac{\bm{v}_i}{\| \bm{v}_i \|_2} - \frac{\bm{u}_n}{\| \bm{u}_n \|_2} \right\|^2
\end{align}
The resulting attention scores are,
\begin{align}
    \operatorname{att}_{n,i}^{(3)} \propto \exp \left( - \kappa \left\| \frac{\bm{v}_i}{\| \bm{v}_i \|_2} - \frac{\bm{u}_n}{\| \bm{u}_n \|_2} \right\|^2 + 2 \kappa Z_i Z_n \right)
\end{align}
Recall that $\mathcal{I}_n = \{ k +1 \le i \le n : \forall j \in [k],\ x_{n-j+1} = x_{i-j} \}$. Then for any $i \in \mathcal{I}_n$, $\bm{v}_i = \bm{u}_n$, and by \Cref{lemma:normerror}, for any $i \ge k+1$ but not in $\mathcal{I}_n$,
\begin{align} \nonumber
    \left\| \frac{\bm{v}_i}{\| \bm{v}_i \|_2} - \frac{\bm{u}_n}{\| \bm{u}_n \|_2} \right\|_2 \ge \frac{1}{3^k}.
\end{align}
Note that this gap is small but non-zero. Furthermore, recall that $Z_i = 3^{k+1}/5$ if $i \ge k$ and otherwise $Z_i \le 3^k/5$. Thus the attention prefers values of $i$ such that $\bm{v}_i = \bm{u}_n$ and such that $i \ge k+1$. In particular, as $\kappa \to \infty$, the resulting attention pattern is,
\begin{align}
    \operatorname{att}^{(3)}_{n,\cdot} &= \operatorname{Unif} ( \mathcal{I}_n ).
\end{align}
Choosing, 
\begin{align} \nonumber
    \bm{W}_V^{(3)} &= \begin{bmatrix}
        \bm{0} & \bm{0} & \bm{0} \\
        \bm{0}_{S \times 3} & I_{S \times S} & \bm{0}
    \end{bmatrix}.
\end{align}
We get that,
\begin{align} \nonumber
    \widetilde{\bm{x}}_n^{(3)} = \bm{x}_n^{(3)} + \sum_{i=1}^n \operatorname{att}_{n,i}^{(3)} \begin{bmatrix}
        \bm{0} \\ e_{x_i}^S
    \end{bmatrix} = \bm{x}_n^{(3)} + \frac{1}{|\mathcal{I}_n|} \sum_{i \in \mathcal{I}_n} \begin{bmatrix}
        \bm{0} \\ e_{x_i}^S
    \end{bmatrix}.
\end{align}
The feedforward layer is chosen to have $\bm{W}_1^{(3)} = \bm{W}_2^{(3)} = \bm{0}$, and the overall output of the final transformer layer is therefore just $\widetilde{\bm{x}}_n^{(3)}$. In the output linear layer, choose,
\begin{align}
    \begin{split}
    \bm{A} &= \begin{bmatrix}
        \bm{0}_{S \times (d-S)} & I_{S \times S}
    \end{bmatrix} \\
    \bm{b} &= \bm{0}
    \end{split}
\end{align}
which results in,
\begin{align}
    \operatorname{logit}_n &= \frac{1}{|\mathcal{I}_n|} \sum_{i \in \mathcal{I}_n} e_{x_i}^S = \sum_{i=k+1}^n \frac{\mathbb{I} ( \forall 1 \le j \le k,\ x_{i-j} = x_{n-j+1})}{\sum_{i' = k+1}^n \mathbb{I} ( \forall 1 \le j \le k,\ x_{i'-j} = x_{n-j+1}) } \cdot e_{x_i} \nonumber
\end{align}
In particular,
\begin{align}
    \operatorname{logit}_T (x_{T+1}) &= \frac{\sum_{n=k+1}^T \mathbb{I} ( \forall 0 \le i \le k,\ x_{n-i} = x_{T-i+1})}{\sum_{n = k+1}^T \mathbb{I} ( \forall 1 \le i \le k,\ x_{n-i} = x_{T-i+1}) }
\end{align}
which is the conditional $k$-gram.

\newpage

\section{Representation lower bounds for $1$-layer transformers: Proof of \Cref{theorem:lb-1layer}}

We prove this lower bound by a reduction to communication complexity, and specifically to the set disjointness problem.

Suppose Alice and Bob are given strings $\bm{a}, \bm{b} \in \{ 0, 1 \}^n$ which are indicator vectors of sets $A$ and $B$. Their goal is to jointly compute $\operatorname{DIS} (\bm{a}, \bm{b}) = \mathbb{I} ( \exists i : \bm{a}_i = \bm{b}_i = 1)$, which indicates whether $A$ and $B$ intersect or not. Alice and Bob may send a single bit message to the other party over a sequence of communication rounds. The following seminal result by \cite{yao1979some} asserts a lower bound on amount of communication required between Alice and Bob to carry out this task. 

\begin{theorem}[\cite{yao1979some}] \label{theorem:yao-lb}
Any deterministic protocol for computing $\operatorname{DIS} (\bm{a}, \bm{b})$ requires at least n rounds of communication.
\end{theorem}

We show that a $1$-layer transformer with sufficiently small embedding dimension / number of heads can be used to simulate a two-way communication protocol between Alice and Bob to solve $\operatorname{DIS} (\bm{a}, \bm{b})$ in a way which contradicts Yao's lower bound in \Cref{theorem:yao-lb}.

With $m = T/3-1$, suppose Alice and Bob have length $m$ bit strings $\bm{a}, \bm{b} \in \{ 0,1 \}^m$. The transformer's input will be a sequence of the form,
\begin{align}
    2, \bm{a}_1 , \bm{b}_1, 2, \bm{a}_2, \bm{b}_2, \cdots, 2, \bm{a}_m, \bm{b}_m, 2, 1,
\end{align}
of length $3m+2 = T-1$. The input basically contains a repeating motif, composed of the symbol $2$ followed by one of Alice's bits, and then one of Bob's bits. The last $2$ symbols are $2$ and $1$. We will consider the empirical conditional $3$-gram probability the transformer associates with the symbol $x_T = 2$. Noting that $x_{T-1} = 1$ and $x_{T-2} = 1$, the conditional $3$-gram is computed to be,
\begin{align}
    \frac{\sum_{i=3}^{T-1} \mathbb{I} (x_i = 1, x_{i-1} = 1, x_{i-2} = 2)}{\sum_{i=3}^{T-1} \mathbb{I} (x_{i-1} = 1, x_{i-2} = 2)}
\end{align}
Note that if $x_{i-2} = 2$, then $i$ must be of the form $3j$ for $j=1,\cdots,n$, and we may rewrite the sum as,
\begin{align} \label{eq:intersection}
    \frac{\sum_{j=1}^m \mathbb{I} (x_{3j} = 1, x_{3j-1} = 1)}{\sum_{j=1}^m \mathbb{I} (x_{3j-1} = 1)} = \frac{|A \cap B|}{|B|}
\end{align}

Now, let us use the transformer to construct a deterministic communication protocol between Alice and Bob. Alice is given $(x_2,x_5,\cdots,x_{3m-1}) = (\bm{a}_1,\bm{a}_2,\cdots,\bm{a}_m)$ and Bob is given $(x_3,x_6,\cdots,x_{3m}) = (\bm{b}_1,\bm{b}_2,\cdots,\bm{b}_m)$.

In the first round, Alice computes the normalization in the softmax of the attention which comes from the set of inputs she holds. For simplifying notation define,
\begin{align}
    \operatorname{score}^{(h)} (i) = \exp \left( \left\langle \bm{W}_K^{(h)} (\texttt{Emb} (x_{i}) + \bm{p}_{i}), \bm{W}_Q^{(h)} \texttt{Emb} (x_{T-1}) \right\rangle \right)
\end{align}

In particular, for each head $h \in [H]$, she computes,
\begin{align}
    Z_{\text{Alice}}^{(h)} = \log \left( \sum\nolimits_{j=1}^m \operatorname{score}^{(h)} (3j-1) \right)
\end{align}
Assuming that the transformer uses $p$ bits of precision, Alice communicates $Z_{\text{Alice}}^{(h)}$ for each $h$, which corresponds to $pH$ bits of communication.
With this information, Bob completes the rest of the normalization term (again up to $p$ bits of precision) and computes,
\begin{align}
    Z^{(h)} &= \log \left( Z^{(h)}_{\text{Alice}} + Z^{(h)}_{\text{Bob}} + Z^{(h)}_{\text{common}} \right), \\
    &\text{where } Z^{(h)}_{\text{Bob}} = \log \left( \sum\nolimits_{j=1}^m \operatorname{score}^{(h)} (3j) \right) \\
    &\text{and } Z^{(h)}_{\text{common}} = \log \left( \sum\nolimits_{j=1}^m \operatorname{score}^{(h)} (3j-2) + \operatorname{score}^{(h)} (T-2) + \operatorname{score}^{(h)} (T-1) \right)
\end{align}
which is the overall normalization term in the softmax. This is communicated back to Alice, using another $pH$ bits of communication. Next using this information, Alice computes the output of the attention layer, taking the convex combination corresponding to the inputs she knows. In particular, for each $h \in [H]$ she computes,
\begin{align}
    \sum_{j=1}^n \frac{\operatorname{score}^{(h)} (3j-1)}{\exp (Z^{(h)})} \texttt{Emb} (x_{3j-1}) \in \mathbb{R}^d.
\end{align}
across all the heads. Rather than transmitting everything, she concatenates the outputs of the heads, and multiplies them by the value and projection matrices to result in the output $\bm{y}_{\text{Alice}}$ which is $d$-dimensional. This is sent to Bob using $d p$ bits of communication. Subsequently, Bob computes the terms in the attention corresponding to the inputs he knows as well as the public inputs (all the $2$'s at positions $3j-2$ as well as the last two symbols). In particular,
\begin{align}
    &\sum_{j=1}^m \frac{\operatorname{score}^{(h)} (3j)}{\exp (Z^{(h)})} \texttt{Emb} (x_{3j}) + \sum_{j=1}^{m+1} \frac{\operatorname{score}^{(h)} (3j-2)}{\exp (Z^{(h)})} \texttt{Emb} (2) + \frac{\operatorname{score}^{(h)} (T-1)}{\exp (Z^{(h)})} \texttt{Emb} (1)
\end{align}
These are yet again concatenated across all the heads and multiplied by the value and projection matrices to result in the output $\bm{y}_{\text{Bob}}$ which is added to $\bm{y}_{\text{Alice}}$ to result in $\bm{y}$. Bob passes $\bm{y}$ through the residual connection, layer norm, and feedforward layers, and subsequently through the linear layer and softmax of the model to result in the output of the model. By assumption, the output of the model approximately captures the conditional $3$-gram, which by \cref{eq:intersection} equals $|A \cap B| / |B|$. Note that if $|A \cap B| / |B|$ is non-zero, it must be at least $1/T$. This means, if the transformer is able to compute the conditional $3$-gram to within an additive error of $1/3T$, then Bob can simply threshold the output of the transformer to decide whether $A \cap B = \emptyset$ or not, thereby solving $\operatorname{DIS} (\bm{a}, \bm{b})$.

Since this communication protocol is deterministic, by Yao's lower bound in \cref{theorem:yao-lb}, the number of bits communicated between Alice and Bob must be at least $m = T/3-1$. The total number of bits of communication in the protocol is $2 pH + d p + 1$ (the last $1$ comes from Bob having to communicate the answer to Alice), completing the proof.

\section{Lower bounds on representing \kth induction heads: Proof of \Cref{theorem:lb-Llayer,theorem:lb-kheads,theorem:lb-general}}

In this section we prove a size-lower bound on attention-only transformers representing $k^{\text{th}}$-order induction heads which generalizes both \Cref{theorem:lb-Llayer} and \Cref{theorem:lb-kheads}. We will first prove the result for the case $L=2$ and $H=1$, showing that they cannot represent $k^{\text{th}}$-order induction heads for $k \ge 4$ under \Cref{ass:1}. We subsequently extend it to the general $L$-layer transformer in \Cref{subsec:LlayerLB} and to the general case with $H_{\ell}$ heads in layer $\ell \in [L]$ in \Cref{subsec:kheadsLB}.

\subsection{Lower bounds on $2$-layer $1$-head attention-only transformers}

In this section we show that under \Cref{ass:1}, a $2$-layer $1$-head attention-only transformer cannot represent \kth induction heads for any $k \ge 4$. We will prove lower bounds on the transformer when the input is binary, i.e., $S = \{ 0,1 \}$. With relative position embeddings, observe that the first layer of the transformer model learns representations of the form,
\begin{align}
    \bm{x}_n^{(2)} = \texttt{Emb} (x_n) + \sum_{i \le n} \operatorname{att}^{(1)}_{n,i} \bm{W}_V^{(1)} \texttt{Emb} (x_{i}) + \sum_{i \le n} \bm{W}_V^{(1)} \bm{p}_{n-i}^{V,(1)}
\end{align}
where note that the attention pattern only depends on $n$ and $i$ and not on $x_i$ or $x_n$. These representations are input into the second layer, which realizes the attention pattern $\operatorname{att}_{n,i}^{(2)}$, which is proportional to,
\begin{align}
    \exp \left( \left\langle \bm{W}_K^{(2)} \big( \bm{x}^{(2)}_i + \bm{p}_{n-i}^{K,(2)} \big), \bm{W}_Q^{(2)} \bm{x}^{(2)}_n \right\rangle \right).
\end{align}
We need this function to be maximized uniquely when $x_{i-1} = x_n, \cdots, x_{i-k} = x_{n-k+1}$. Denoting $\phi(0) = \bm{W}_V^{(1)} \texttt{Emb} (0)$ and $\phi (1) = \bm{W}_V^{(1)} \texttt{Emb} (1)$,
\begin{align}
    \bm{x}_n^{(2)} &= \texttt{Emb} (x_n) + \sum_{i \le n} \operatorname{att}^{(1)}_{n,i} \bm{W}_V^{(1)} \texttt{Emb} (x_i) + \sum_{i \le n} \bm{W}_V^{(1)} \bm{p}_{n-i}^{(1),V}\\
    &= x_n \texttt{Emb} (1) + (1-x_n) \texttt{Emb} (0) + \sum_{i \le n} \operatorname{att}^{(1)}_{n,i} \big( x_i \cdot \phi(1) + (1-x_i) \cdot \phi(0) \big) + \sum_{i \le n} \bm{W}_V^{(1)} \bm{p}_{n-i}^{(1),V} \\
    &= \left( \frac{\texttt{Emb} (1) + \texttt{Emb} (0)}{2} + x_n' \cdot \frac{\texttt{Emb} (1) - \texttt{Emb} (0)}{2} \right) + \sum_{i \le n} \operatorname{att}^{(1)}_{n,i} \left( \frac{\phi(1)+\phi(0)}{2} + x_i' \cdot \frac{\phi(1)-\phi(0)}{2} \right) \nonumber \\
    &\hspace{25em} +\sum_{i \le n} \bm{W}_V^{(1)} \bm{p}_{n-i}^{(1),V}
\end{align}
where $x_i' \gets 2x_i - 1$. We can write this down as,
\begin{align} \label{eq:simple}
    \bm{x}_n^{(2)} = \bm{m}_n^{(1)} + \bm{M}_n^{(1)} \begin{bmatrix}
        x_n' & x_{n-1}' & \cdots & x_1'
    \end{bmatrix}^T
\end{align}
where $\bm{M}_n^{(1)}$ is a matrix of rank at most $2$ and of the form,
\begin{align} \label{eq:Mn1}
    \bm{M}_n^{(1)} = \left( \frac{\phi (1) - \phi(0)}{2} \right) \begin{bmatrix}
    \operatorname{att}^{(1)}_{n,n} & \cdots & \operatorname{att}^{(1)}_{n,1}
\end{bmatrix} + \left( \frac{\texttt{Emb} (1) - \texttt{Emb} (0)}{2} \right) \begin{bmatrix}
    1 & 0 & \cdots & 0
\end{bmatrix}
\end{align}
which is independent of $x_1',\cdots,x_n'$. Likewise $\bm{m}_n^{(1)}$ collects all the vectors in the sum that don't depend on $x_1',\cdots,x_n'$.
Now, observe that in the next layer, we wish to show that an induction head cannot be realized by $\operatorname{att}_{n,i}^{(2)}$ for each $i \le n$. We will show this for any value of $i \le n-k$.

In the second layer, we may write down the key vectors as,
\begin{align}
    \bm{W}_K^{(2)} \left( \bm{x}_i^{(2)} + \bm{p}_{n-i}^{(2),K} \right) &= \bm{W}_K^{(2)} \bm{m}_i^{(1)} + \bm{W}_K^{(2)} \bm{M}_i^{(1)} \begin{bmatrix}
        x_i' & x_{i-1}' & \cdots & x_1'
    \end{bmatrix}^T + \bm{W}_K^{(2)} \bm{p}_{n-i}^{(2),K}.
\end{align}
Again, defining the vector $\overline{\bm{m}}_i^{(1)}$ and the matrix $\overline{\bm{M}}_i^{(1)}$ appropriately (having rank at most $2$), this equals,
\begin{align}
    \overline{\bm{m}}_i^{(1)} ( \{x_i' \} \cup \{ x_{-k-1}' ,\cdots,x_1'\}) + \overline{\bm{M}}_i^{(1)} \bm{y}
\end{align}
where $\bm{y} \triangleq \begin{bmatrix}
        x_{i-1}' & \cdots & x_{i-k}'
    \end{bmatrix}^T$ and the vector $\overline{\bm{m}}_i^{(1)}$ depends on $x_i'$ as well as the inputs $x_{i-k-1}',\cdots,x_1'$, which in this context, are treated as nuisance variables since they do not intersect with $\{ x_{i-1}',\cdots, x_{i-k}'\} \cup \{ x_n , \cdots, x_{n-k+1} \}$. Henceforth we will avoid explicitly stating the dependency of $\overline{\bm{m}}_i^{(1)}$ on the $x_j$'s. Similarly, the query vector can be written down as,
\begin{align}
    \bm{W}_Q^{(2)} \bm{x}_n^{(2)} = \bmhat{m}_n^{(1)} + \widecheck{\bm{M}}_n^{(1)} \bm{x} + \bmhat{M}_n^{(1)} \bm{y}
\end{align}
where $\bmhat{m}_n^{(1)}$, $\widecheck{\bm{M}}_n^{(1)}$ and $\bmhat{M}_n^{(1)}$ are defined appropriately, with $\bmhat{M}_n^{(1)}$ and $\bmhat{M}_n^{(1)}$ of rank at most $2$, and $\bm{x}$ is defined as $\begin{bmatrix}
        x'_{n} & \cdots & x'_{n-k+1}
    \end{bmatrix}^T$. For an appropriate matrix $\bm{M}^{\times}_{n,i}$, vectors $\bm{m}^\times_{n,i}$ and $\widetilde{\bm{m}}^{\times}_{n,i}$ and scalar $m_{n,i}^{\times}$, the dot-product of the key and query vectors can be written as,
\begin{align} \label{eq:bilinear}
    &\left\langle \bm{W}_K^{(2)} \big( \bm{x}^{(2)}_i + \bm{p}_{n-i}^{(2),K} \big), \bm{W}_Q^{(2)} \bm{x}^{(2)}_n \right\rangle \nonumber\\
    &\quad = \bm{x}^T \bm{M}_{n,i}^\times \bm{y} + \bm{y}^T \big( \overline{\bm{M}}_{n,i}^{\times} \big) \bm{y} + (\bm{m}_{n,i}^\times)^T \bm{x} + (\overline{\bm{m}}^{\times}_{n,i})^T \bm{y} + m_{n,i}^\times \triangleq f_{n,i} (\bm{x}, \bm{y}),
\end{align}
Which is a linear function in $\bm{x}$ and quadratic in $\bm{y}$, both of which lie on $\{ \pm 1 \}^k$. Note that the matrix $\bm{M}_{n,i}^\times$ has rank at most $2$ since it is a product of $\overline{\bm{M}}_i^{(1)}$ and $\widecheck{\bm{M}}_n^{(1)}$, each with rank at most $2$. Next we introduce a lemma showing that if $\bm{M}_{n,i}^\times$ is inherently low rank, the quadratic form in \cref{eq:bilinear} which captures the dot-product between the key and value vectors cannot satisfy the property that for every $\bm{y}$, the function is uniquely maximized at $\bm{x} = \bm{y}$. In particular, this means that for any $i \le n-k$, there is some choice of $x_n,x_{n-1},\cdots,x_{n-k+1}$ such that there are $x_{i-1},\cdots,x_{i-k}$ such that for at least one $j \in [k]$, $x_{i-j}$ and $x_{n-j-1}$ are not equal, but the attention score is larger than the case when $x_{i-j}$ were equal to $x_{n-j-1}$ for each $j \in [k]$.

\begin{lemma} \label{lemma:lowrank}
If $\bm{M}_{n,i}^\times$ has rank $\le k-2$, it is impossible for $f_{n,i} (\bm{x}, \bm{y})$ to satisfy the property that for every $\bm{y} \in \{ \pm 1 \}^k$, the maximizer is uniquely $\bm{x} = \bm{y}$.
\end{lemma}

The proof is almost complete: if $k \ge 4$, then the rank of $\bm{M}_{n,i}^\times$, which is at most $2$, does not exceed $k-2$. This means that when $k \ge 4$, any attention pattern realized in the second layer must satisfy the property that there exists a string such that the attention is no longer uniquely maximized when $x_n = x_{i-1}, \cdots, x_{n-k+1} = x_{i-k}$.

\begin{proof}
For the purpose of brevity, define $\mathcal{H}_k = \{ \pm 1 \}^k$. First consider the reparameterization,
\begin{equation}
    \widetilde{\bm{x}} = \widetilde{\bm{M}}_{n,i}^\times \bm{x}, \text{ where } \widetilde{\bm{M}}_{n,i}^\times = \begin{bmatrix} (\bm{M}_{n,i}^\times)^T \\ (\bm{m}_{n,i}^\times)^T \end{bmatrix}.
\end{equation}
Then, the dot-product of the key and query matrices can be written as,
\begin{align} \label{eq:obj71}
    \begin{bmatrix} \bm{y}^T & 1 \end{bmatrix} \widetilde{\bm{x}} + \bm{y}^T \big( \overline{\bm{M}}_{n,i}^{\times} \big) \bm{y} + (\overline{\bm{m}}_{n,i}^\times)^T \bm{y} + m_{n,i}^\times
\end{align}
Note that this function is linear in $\widetilde{\bm{x}}$ and therefore must be maximized on a vertex of the convex hull of the domain, $\widetilde{\bm{M}}_{n,i}^\times \mathcal{H}_k \triangleq \big\{ \widetilde{\bm{M}}_{n,i}^\times \bm{h} : \bm{h} \in \mathcal{H}_k \big\}$. If $\bm{M}_{n,i}^\times$ has rank at most $k-2$, the rank of $\widetilde{\bm{M}}_{n,i}^\times$ is at most $k-1$ and cannot be full rank. We show that this must imply that there is a vertex $\bm{v} \in \mathcal{H}_k$ such that $\widetilde{\bm{M}}_{n,i}^\times \bm{v}$ is not a unique vertex of the convex hull of $\widetilde{\bm{M}}_{n,i}^\times \mathcal{H}_k$. This means that $\bm{v}$ cannot be a unique maximizer for $\widetilde{\bm{x}}$ when maximizing over all strings in \cref{eq:obj71}, and specifically $\bm{y} = \bm{v}$ is a witness to \Cref{lemma:lowrank}.

Below we discuss how to find such a vector $\bm{v}$. Note that $\widetilde{\bm{M}}_{n,i}^\times$ is not full rank, which implies that there exists a vector $\bm{n}$ such that $\widetilde{\bm{M}}_{n,i}^\times \bm{n} = \bm{0}$. Without loss of generality, let $\bm{n}_1$ be the smallest non-zero coordinate of $\bm{n}$ in absolute value. Then the vector $\bm{n}_1^{-1} \bm{n}$ has no non-zero coordinates in the interval $(-1,1)$. We will show that $\operatorname{sign} (\bm{n}_1^{-1} \bm{n})$ is a good choice for $\bm{v}$.

Consider two cases,

\paragraph{Case I.} Every non-zero coordinate of $\bm{n}_1^{-1} \bm{n}$ is in $\{ \pm 1 \}$.
Consider any $\bm{x} \in \mathcal{H}_k$ which matches with $\bm{n}$ on the non-zero coordinates. Consider $\bm{x}'$ which is the same as $\bm{x}$, except a negation is taken on the coordinates where $\bm{n}$ is non-zero. Note that $\widetilde{\bm{M}}_{n,i}^\times \bm{x} = \widetilde{\bm{M}}_{n,i}^\times \bm{x}'$, for the same value of $\bm{x}$. This means that for any $\bm{y}$. In particular, from \cref{eq:obj71}, both $\bm{x}$ and $\bm{x}'$ are maximizers, showing that \Cref{lemma:lowrank} is true in this case. We circumvent having to find such a vector $\bm{v}$ in this case.

\paragraph{Case II.} $\bm{n}_1^{-1} \bm{n}$ has non-zero coordinates which are not all in $\{ \pm 1 \}$. In particular, at least one coordinate where this vector is strictly less than $-1$ or strictly greater than $+1$.
In this case, observe that the sign vector $\widetilde{\bm{n}} = \operatorname{sign} (\bm{n}_1^{-1} \bm{n}) \in \mathcal{H}_k$ lies within, but is not a vertex of the convex hull of the set $\mathcal{H}_k \cup \{ \bm{n}_1^{-1} \bm{n} \}$. The reason for this is simple to see when we assume that $\bm{n}_1^{-1} \bm{n}$ has only one coordinate which is not in $[-1,1]$, say, the coordinate $j=2$: here, $\widetilde{\bm{n}}$ can be written down as a convex combination (with non-zero coefficients) of $\bm{n}_1^{-1} \bm{n}$ and $\widetilde{\bm{n}}^{(2)}$; the latter vector is obtained by flipping coordinate $2$ of $\widetilde{\bm{n}}$. When there is more than one coordinate not in $[-1,1]$, we can peel away these large coordinates in $\bm{n}_1^{-1} \bm{n}$ by taking a convex combination of this vector with the vectors $\widetilde{\bm{n}}^{(j)}$ for the appropriate values of $j$, to return the sign vector $\widetilde{\bm{n}}$. Here, $\widetilde{\bm{n}}^{(j)}$ is the version of $\widetilde{\bm{n}}$ where the $j^{\text{th}}$-coordinate is flipped. This results in the following claim.

\begin{claim}
The sign vector $\widetilde{\bm{n}}$ lies within the convex hull of the points $\mathcal{H}_k \cup \{ \bm{n}_1^{-1} \bm{n} \}$, but is not a vertex of this set.
\end{claim}

In particular, we may write,
\begin{align}
    \widetilde{\bm{n}} = \alpha_0 \bm{n}_1^{-1} \bm{n} + \sum_{j \in [n]} \alpha_j \widetilde{\bm{n}}^{(j)}.
\end{align}
where $\alpha_0 > 0$ and $\sum_{j=0}^n \alpha_i = 1$. By left-multiplying this on both sides by $\widetilde{\bm{M}}_{n,i}^\times$ and noting that $\bm{n}$ lies in the null-space of this matrix, we get,
\begin{align}
    \widetilde{\bm{M}}_{n,i}^\times \widetilde{n} = \sum_{j \in [n]} \alpha_j \widetilde{\bm{M}}_{n,i}^\times \widetilde{\bm{n}}^{(j)}
\end{align}
where note that $\sum_{j \in [n]} \alpha_j$ is strictly less than $1$, since $\alpha_0 > 0$. We may write this vector as,
\begin{align}
    \widetilde{\bm{M}}_{n,i}^\times \widetilde{n} &= \alpha_0 \bm{0} + \sum_{j \in [n]} \alpha_j \widetilde{\bm{M}}_{n,i}^\times \widetilde{\bm{n}}^{(j)} \nonumber\\
    &= \frac{\alpha_0}{2^k} \sum_{\bm{h} \in \mathcal{H}_k} \widetilde{\bm{M}}_{n,i}^\times \bm{h} + \sum_{j \in [n]} \alpha_j \widetilde{\bm{M}}_{n,i}^\times \widetilde{\bm{n}}^{(j)}
\end{align}
Since $\alpha_0 > 0$, this equation implies that the image of $\widetilde{\bm{n}}$ under $\widetilde{\bm{M}}_{n,i}^\times$ itself falls within  $\operatorname{conv} (\widetilde{\bm{M}}_{n,i}^\times \mathcal{H}_k)$, but is itself not a vertex of this set.
This means that $\widetilde{\bm{n}}$ can never be a maximizer of $f_{n,i} \big( \cdot,\bm{y} \big)$ for any $\bm{y}$, and in particular when $\bm{y} = \widetilde{\bm{n}}$, thereby proving \Cref{lemma:lowrank}.
\end{proof}

\subsection{$L$-layer attention-only transformers with $1$ head per layer: Proof of \Cref{theorem:lb-Llayer}} 
\label{subsec:LlayerLB}

\begin{proof}
The proof largely tracks the $2$-layer case, with the main exception that we keep track of how the maximum possible rank of the matrix $\bm{M}_{n,i}^\times$ grows as a function of the depth of the transformer. In the case the $2$-layer transformer, we show that it cannot exceed $2$. With the addition of more layers, we show that it cannot exceed $2^{L-1}$.

Recall from the notation in \cref{eq:simple} that the output of the first attention layer is,
\begin{align}
\bm{x}_n^{(2)}
&= \bm{m}^{(1)}_n+ \bm{M}^{(1)}_n \begin{bmatrix}
    x_n' & x_{n-1}' & \cdots & x_1'
\end{bmatrix}^T
\end{align}
where $\bm{M}^{(1)}_n \in \mathbb{R}^{d \times n}$ has rank at most $2$. Let us rewrite this as,
\begin{align}
    \bm{x}_n^{(2)}
&= \bm{m}^{(1)}_n+ \overline{\bm{M}}^{(1)}_n \begin{bmatrix}
    x_T' & x_{T-1}' & \cdots & x_1'
\end{bmatrix}^T
\end{align}
where $\bm{M}^{(1)}_n \in \mathbb{R}^{d \times T}$ is causally masked to be $0$'s when it operates on $x_i$ for all indices $i > n$. Note that even with this causal masking, $\overline{\bm{M}}_n^{(1)}$ has rank at most $2$, as discussed in \cref{eq:simple}.

By induction, assume that the output of the $(\ell-1)^{\text{th}}$ attention layer is of the form,
\begin{align}
    \bm{x}_n^{(\ell)} = \bm{m}_n^{(\ell-1)} + \overline{\bm{M}}_n^{(\ell-1)}\bm{x}_{1:T}
\end{align}
where $\bm{x}_{1:T} \triangleq  \begin{bmatrix}
    x_T' & x_{T-1}' & \cdots & x_1'
\end{bmatrix}^T$. Passing $\bm{x}_n^{(\ell)}$ through the $\ell^{\text{th}}$ attention layer, we get,
\begin{align}
    \bm{x}_n^{(\ell+1)} &= \bm{x}_n^{(\ell)} + \sum_{i \le n} \operatorname{att}^{(\ell)}_{n,i} \bm{W}_V^{(\ell)} \left( \bm{x}_i^{(\ell)} + \bm{p}_{n-i}^{(\ell),V} \right) \\ 
    &= \bm{m}_n^{(\ell-1)} + \overline{\bm{M}}_n^{(\ell-1)} \bm{x}_{1:T} + \sum_{i \le n} \operatorname{att}_{n,i}^{(\ell)} \bm{W}_V^{(\ell)} \bm{m}_i^{(\ell-1)} + \sum_{i \le n} \operatorname{att}_{n,i}^{(\ell)} \bm{W}_V^{(\ell)} \overline{\bm{M}}_i^{(\ell-1)} \bm{x}_{1:T} \nonumber\\
    &\hspace{22em}+ \sum_{i \le n} \operatorname{att}_{n,i}^{(\ell)} \bm{W}_V^{(\ell)} \bm{p}_{n-i}^{(\ell),V}
\end{align}
Define,
\begin{align} \label{eq:mln}
    \bm{m}_n^{(\ell)} &= \bm{m}_n^{(\ell-1)} + \sum_{i \le n} \operatorname{att}_{n,i}^{(\ell)} \bm{W}_V^{(\ell)} \bm{m}_i^{(\ell-1)} + \sum_{i \le n} \operatorname{att}_{n,i}^{(\ell)} \bm{W}_V^{(\ell)} \bm{p}_{n-i}^{(\ell),V}\text{, and,}\\
    \overline{\bm{M}}_n^{(\ell)} &= \overline{\bm{M}}_n^{(\ell-1)} + \sum_{i \le n} \operatorname{att}_{n,i}^{(\ell)} \bm{W}_V^{(\ell)} \overline{\bm{M}}_i^{(\ell-1)}
\end{align}
Then, we can write down,
\begin{align}
    \bm{x}_n^{(\ell+1)} = \bm{m}_n^{(\ell)} + \overline{\bm{M}}_n^{(\ell)} \bm{x}_{1:T}
\end{align}
We also inductively assume that for every $i \le n$,
\begin{enumerate}
    \item[$(i)$] $\overline{\bm{M}}_i^{(\ell-1)}$ has rank $R \le 2^{\ell-1}$, and,
    \item[$(ii)$] $\overline{\bm{M}}_i^{(\ell-1)}$ can be factorized in the form $\sum_{r=1}^R \bm{u}_r \cdot \bm{v}_{i,r}^T$, where only the $\bm{v}_{i,r}$'s depend on $i$, but the $\bm{u}_{r}$'s do not depend on $i$.
\end{enumerate}
Both of these conditions are true when $\ell-1 = 1$ as evidenced by the structure of $\bm{M}_i^{(1)}$ in \cref{eq:Mn1} and noting that $\overline{\bm{M}}_i^{(1)}$ is obtained from $\bm{M}_i^{(1)}$ by right multiplying by a diagonal mask matrix.
Using the recursion in \cref{eq:mln}, we prove that the induction hypotheses $(i)$ and $(ii)$ are true at layer $\ell$ as well. In particular using the decomposition in $(ii)$, observe that,
\begin{align}
    \overline{\bm{M}}_n^{(\ell)} &= \sum_{r=1}^R \bm{u}_r \cdot \bm{v}_{n,r}^T + \sum_{i \le n} \operatorname{att}_{n,i}^{(\ell)} \bm{W}_V^{(\ell)} \sum_{r=1}^R \bm{u}_r \cdot \bm{v}_{i,r}^T \\
    &= \sum_{r=1}^R \bm{u}_r \cdot \bm{v}_{n,r}^T + \sum_{r=1}^R \bm{W}_V^{(\ell)} \bm{u}_r \cdot \left( \sum\nolimits_{i \le n} \operatorname{att}_{n,i}^{(\ell)} \bm{v}_{i,r} \right)^T \\
    &= \sum_{r=1}^{2R} \bm{u}_r \cdot \bm{v}_{n,r}^T
\end{align}
where for $r' \in [R]$, $\bm{u}_{R+r'} \triangleq \bm{W}_V^{(\ell)} \bm{u}_r$ and $\bm{v}_{n,r'} \triangleq \sum\nolimits_{i \le n} \operatorname{att}_{n,i}^{(\ell)} \bm{v}_{i,r}$. Since $\bm{M}_n^{(\ell)}$ is the sum of $2R$ rank $1$ matrices and therefore has rank at most $2R \le 2^{\ell}$, proving both parts of the induction hypothesis.

By induction, at the end of the $(L-1)^{\text{th}}$ layer, we have an output which looks like,
\begin{align}
    \bm{x}_n^{(L)} = \bm{m}_n^{(L-1)} + \overline{\bm{M}}_n^{(L-1)} \bm{x}_{1:T}
\end{align}
where $\bm{M}_n^{(L-1)}$ has rank at most $2^{L-1}$. More importantly, note that by the causal masking, even though it appears to depend on the whole input sequence through $\bm{x}_{1:T}$, note that $\bm{x}_n^{(L)}$ only depends on $x_1,\cdots,x_n$ and not on the future inputs to this time $n$. In particular, by a similar argument as in the $2$-layer case (cf. \cref{eq:simple} to \cref{eq:bilinear}), for any $i \le n-k$ we can decompose the dot-product of the key and query vectors at the $L^{\text{th}}$ layer as a bilinear form which looks like,
\begin{align}
    &\left\langle \bm{W}_K^{(L)} \big( \bm{x}^{(L)}_i + \bm{p}_{n-i}^{(L),K} \big), \bm{W}_Q^{(L)} \bm{x}^{(L)}_n \right\rangle \nonumber\\
    &\qquad\qquad = \bm{x}^T \bm{M}_{n,i}^\times \bm{y} + \bm{y}^T \big( \overline{\bm{M}}_{n,i}^{\times} \big) \bm{y} + (\bm{m}_{n,i}^\times)^T \bm{x} + (\overline{\bm{m}}^{\times}_{n,i})^T \bm{y} + m_{n,i}^\times \triangleq f_{n,i}^{(L+1)} (\bm{x}, \bm{y})
\end{align}
where $\bm{x}$ and $\bm{y}$ are defined as $\begin{bmatrix}
        x_{n}' & \cdots & x_{n-k+1}'
    \end{bmatrix}^T$ and $\begin{bmatrix}
        x_{i-1}' & \cdots & x_{i-k}'
    \end{bmatrix}^T$ respectively, and $\bm{M}_{n,i}^\times$ has rank at most that of $\bm{M}_n^{(L-1)}$, which is $2^{L-1}$. In particular, if $2^{L-1} \le k-2$, by \Cref{lemma:lowrank} the proof concludes.
\end{proof}

\subsection{The general case: Transformers with $H_\ell$ heads in layer $\ell$: Proof of \Cref{theorem:lb-general}} \label{subsec:kheadsLB}

The $h^{\text{th}}$ head of the first layer of the attention-only transformer learns patterns of the form,
\begin{align}
    \widetilde{\bm{x}}_n^{(1,h)} &= \sum_{i \le n} \operatorname{att}^{(1,h)}_{n,i} \bm{W}_V^{(1,h)} \texttt{Emb} (x_{i}) + \sum_{i \le n} \bm{W}_V^{(1,h)} \bm{p}_{n-i}^{V,(1,h)} \\
    &= \sum_{i \le n} \operatorname{att}^{(1,h)}_{n,i} \left( \frac{\phi^h (0)+\phi^h (1)}{2} + x_i' \cdot \frac{\phi^h (1)-\phi^h (0)}{2} \right) + \sum_{i \le n} \bm{W}_V^{(1,h)} \bm{p}_{n-i}^{V,(1,h)}
\end{align}
where the last equation assumes a binary input sequence, defines $x_i' = 2x_i-1$ and uses the notation $\phi^h (0) = \bm{W}_V^{(1,h)} \texttt{Emb} (0)$ and $\phi^h (1) = \bm{W}_V^{(1,h)} \texttt{Emb} (1)$. We can further rewrite this as,
\begin{align}
    \widetilde{\bm{x}}_n^{(1,h)} = \bm{m}_n^{(1,h)} + \bm{M}_n^{(1,h)} \bm{x}_{1:T}
\end{align}
where each $\bm{M}_n^{(1,h)} \in \mathbb{R}^{d \times T}$ is rank $1$ and applies a causal mask on the inputs $x_i$ for $i > n$. Recall that the output of the first attention layer applies a projection matrix on the concatentation of $\widetilde{\bm{x}}_n^{(1,h)}$ across $h \in [H_1]$ and then adds a residual connection. The output can be written down as,
\begin{align}
    \widetilde{\bm{x}}_n^{(2)}
    &= \texttt{Emb} (x_n) + \bm{W}_O^{(1)} \begin{bmatrix}
        \bm{m}_n^{(1,1)} \\ \vdots \\ \bm{m}_n^{(1,H_1)}
    \end{bmatrix} + \bm{W}_O^{(1)} \begin{bmatrix}
        \bm{M}_n^{(1,1)} \\ \vdots \\ \bm{M}_n^{(1,H_1)} \end{bmatrix} \bm{x}_{1:T} \\
    &= \bm{m}_n^{(1)} + \bm{M}_n^{(1)} \bm{x}_{1:T},
\end{align}
where,
\begin{align}
    \bm{M}_n^{(1)} &= \left( \frac{\texttt{Emb} (1) + \texttt{Emb} (0)}{2} \right) e_n^T + \bm{W}_O^{(1)} \begin{bmatrix}
        \bm{M}_n^{(1,1)} \\ \vdots \\ \bm{M}_n^{(1,H_1)} \end{bmatrix}, \text{ and,}\\
    \bm{m}_n^{(1)} &= \left( \frac{\texttt{Emb} (1) - \texttt{Emb} (0)}{2} \right) + \bm{W}_O^{(1)} \begin{bmatrix}
        \bm{m}_n^{(1,1)} \\ \vdots \\ \bm{m}_n^{(1,H_1)}
    \end{bmatrix}
\end{align}
Notice that the rank of the matrix $\bm{M}_n^{(1)}$ is at most $H_1+1$. This is because the concatenation operation can increase the rank at most additively, and since each of the $\bm{M}_n^{(1,h)}$ matrices are rank at most $1$.

Following through the proof in \Cref{subsec:LlayerLB} for the $L$-layer case, we can prove inductively that at any layer $\ell$, the output looks like,
\begin{align}
    \bm{x}_n^{(\ell)} = \bm{m}_n^{(\ell)} + \bm{M}_n^{(\ell)} \bm{x}_{1:T}
\end{align}
where the rank of $\bm{M}_n^{(\ell)}$ is $\prod_{i = 1}^{\ell} (H_i + 1)$. Invoking \Cref{lemma:lowrank}, if $\prod_{i = 1}^{L-1} (H_i + 1) \le k-2$, the attention-only transformer cannot realize a \kth induction head at layer $L$.

\section{Model architecture and hyper-parameters}
\label{app:architecture}

The experiments were run on one $8 \times A100$ GPU node.

\begin{table}[h]
\caption{Parameters in the transformer architecture with their shape.}
\label{tab:transformer-parameters}
\vspace{1mm}
\small%
\newcolumntype{R}{>{\raggedleft\arraybackslash}X}
\begin{tabularx}{\linewidth}{Xllr}
\toprule
Parameter
& Matrix shape \\
\cmidrule(lr){1-2}
transformer.wte 
& $2 \times d$ \\
transformer.wpe 
& $N \times d$ \\
transformer.h.ln\_1 $(\times \ell)$
& $d \times 1$ \\
transformer.h.attn.c\_attn $(\times \ell)$
& $3d \times d$ \\
transformer.h.attn.c\_proj $(\times \ell)$
& $d \times d$ \\
transformer.h.ln\_2 $(\times \ell)$
& $d \times 1$ \\
transformer.h.mlp.c\_fc $(\times \ell)$
& $4d \times d$ \\
transformer.h.mlp.c\_proj $(\times \ell)$
& $d \times 4d$ \\
transformer.ln\_f
& $d \times 1$ \\
\bottomrule
\end{tabularx}
\end{table}

\begin{table}[h]
\caption{Settings and parameters for the transformer model used in the experiments.}
\label{tab:transformer-setup}
\vspace{1mm}
\small%
\newcolumntype{R}{>{\raggedleft\arraybackslash}X}
\begin{tabularx}{\linewidth}{lX}
    \toprule
    Dataset & $k$-th order binary Markov source \\
    Architecture & Based on the \gptwo architecture as implemented in \cite{pagliardini-llm} \\
    \midrule
    Batch size & Grid-searched in $\{8, 16\}$ \\
    Accumulation steps & 1 \\
    \midrule
    Optimizer & AdamW ($\beta_1 = 0.9, \beta_2 = 0.95$) \\
    Learning rate & $0.001$ \\
    Scheduler & Cosine \\
    \# Iterations & Up to $25000$ \\
    Weight decay & $\num{1e-3}$\\
    \midrule
    Dropout & $0$ \\
    Sequence length & Grid-searched in $\{32, 64, 128, 256, 512, 1024\}$ \\
    Embedding dimension & Grid-searched in $\{16,32,64\}$ \\
    Transformer layers & Between $1$ and $8$ \\
    Attention heads & Up to $k$ \\
    \midrule
    Repetitions & 3\\
    \bottomrule
\end{tabularx}
\end{table}

\section{Additional experimental results}
\label{app:justification}

\twocolumn

\begin{figure}[t]
\captionsetup[sub]{font=scriptsize}
    \centering
\begin{subfigure}
\centering
   \includegraphics[width=0.41\textwidth]{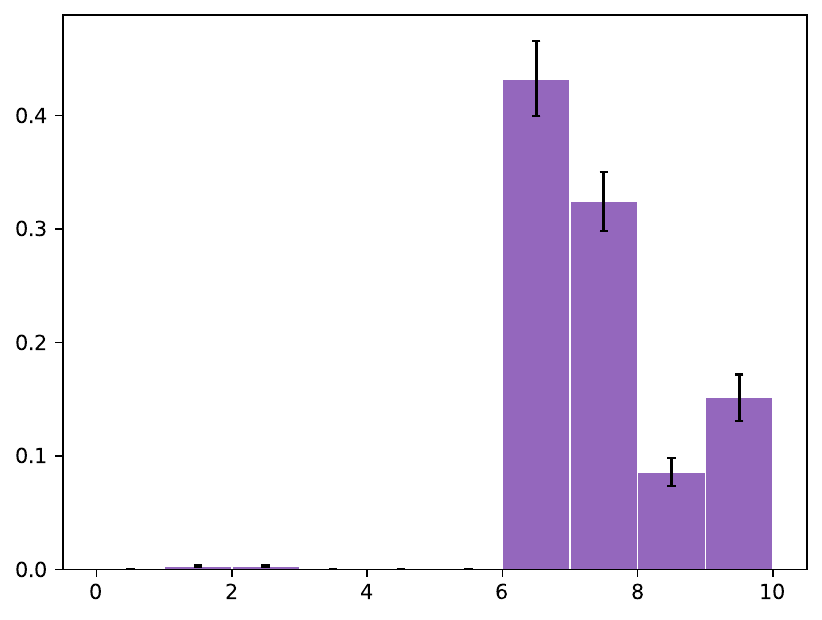} 
   \put(-115,-4){\fontsize{6}{3}\selectfont Sequence index}
      \put(-205,77){\rotatebox[origin=t]{90}{\fontsize{5}{3}\selectfont Attention weight}}
    \caption{First head}
\end{subfigure}
\begin{subfigure}
\centering
   \includegraphics[width=0.41\textwidth]{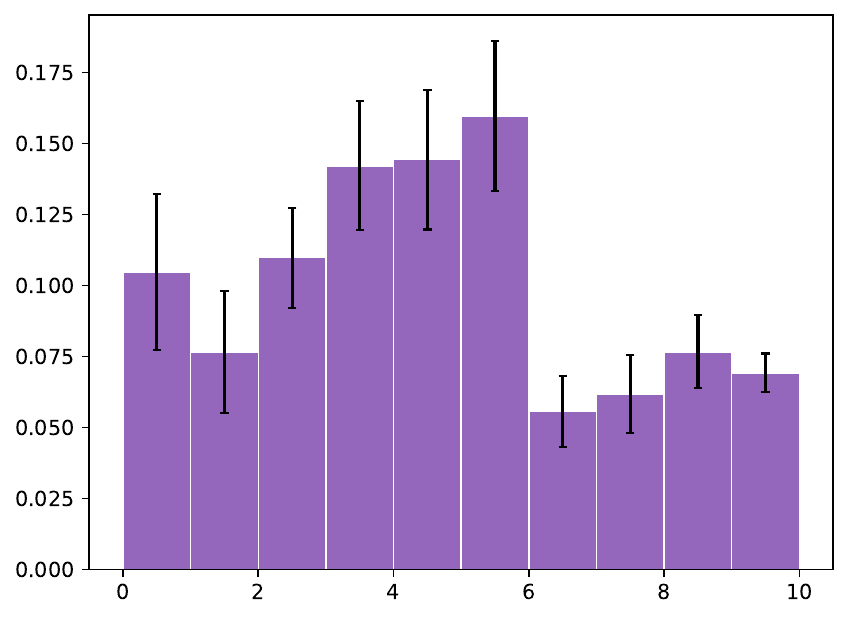} 
   \put(-115,-4){\fontsize{6}{3}\selectfont Sequence index}
      \put(-205,77){\rotatebox[origin=t]{90}{\fontsize{5}{3}\selectfont Attention weight}}
    \caption{Second head}
\end{subfigure}
\caption{Mean attention for column $n=10$ of the two heads of the first attention layer, for a $2$-layer $2$-head transformer model trained on an order-$2$ Markov process, averaged across $100$ input sequences of length $128$.}
\label{fig:hist}
\end{figure}

\begin{figure}[t]
\captionsetup[sub]{font=scriptsize}
    \centering
\begin{subfigure}
\centering
   \includegraphics[width=0.41\textwidth]{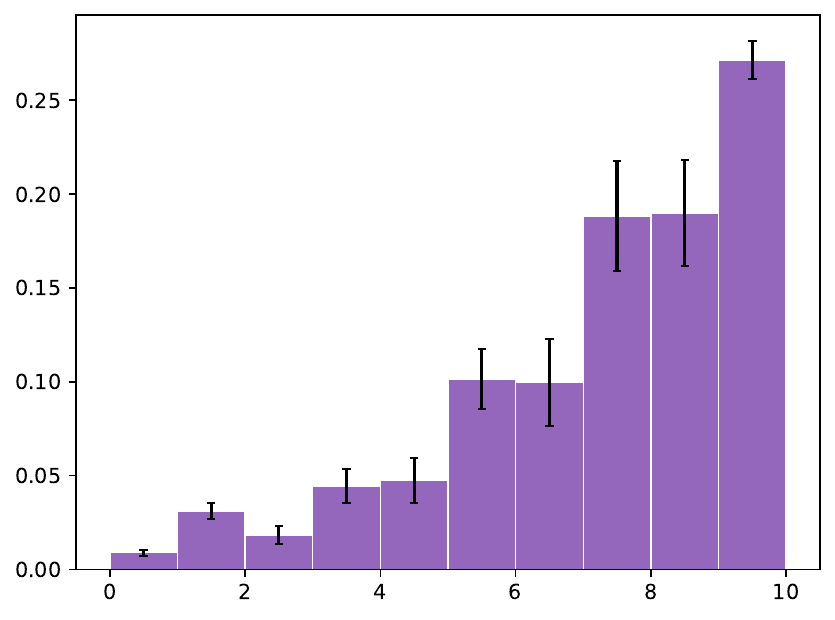} 
   \put(-115,-4){\fontsize{6}{3}\selectfont Sequence index}
      \put(-205,77){\rotatebox[origin=t]{90}{\fontsize{5}{3}\selectfont Attention weight}}
    \caption{First head}
\end{subfigure}
\begin{subfigure}
\centering
   \includegraphics[width=0.41\textwidth]{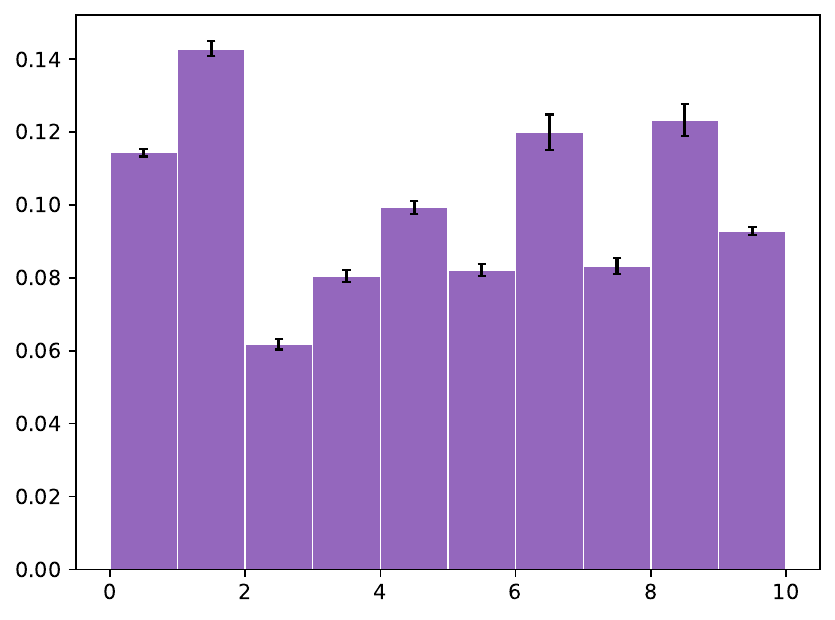} 
   \put(-115,-4){\fontsize{6}{3}\selectfont Sequence index}
      \put(-205,77){\rotatebox[origin=t]{90}{\fontsize{5}{3}\selectfont Attention weight}}
    \caption{Second head}
\end{subfigure}
\begin{subfigure}
\centering
   \includegraphics[width=0.41\textwidth]{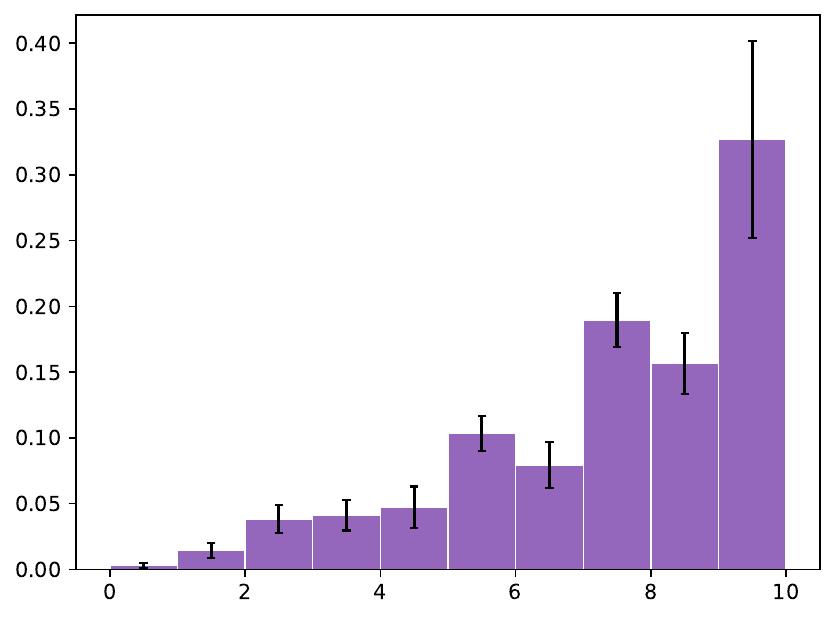} 
   \put(-115,-4){\fontsize{6}{3}\selectfont Sequence index}
      \put(-205,77){\rotatebox[origin=t]{90}{\fontsize{5}{3}\selectfont Attention weight}}
    \caption{Third head}
\end{subfigure}
\caption{Mean attention for column $n=10$ of the three heads of the first attention layer, for a $2$-layer $3$-head transformer model trained on an order-$3$ Markov process, averaged across $100$ input sequences of length $128$.}
\label{fig:hist-2}
\end{figure}

\onecolumn
\Cref{ass:1} suggests that the attention patterns $\operatorname{att}_{n,i}^{(\ell)}$ in layers $\ell = 1,2,\cdots,L-1$, as learnt by an $L$-layer attention-only transformers may only be a function of only the position indices $n,i$. In this section we run some additional experiments to test this conjecture. We train a $2$ layer attention-only transformer with $k$ heads in the first layer, on data drawn from a randomly sampled \kth Markov process, and focus on the learnt attention patterns as a function of in the input sequence. \Cref{fig:hist} plots the results of this experiment for $k=2$ and \Cref{fig:hist-2} for $k=3$. While in both cases there is some variance in the attention patterns learnt by the transformer in some of the heads, we believe that this is a consequence of the iteration budget of the transformer, and specifically the fact that even if the test loss appears to have converged, the transformer may still continue changing in the parameter space. Furthermore, when the attention patterns have some non-zero but small variance as a function of the input, a relaxation of \Cref{ass:1}, we also believe that the results we proved in \Cref{theorem:lb-kheads,theorem:lb-Llayer,theorem:lb-general} should carry over approximately and leave this as an interesting question for future work. Conditional lower bounds of this nature, reliant on structural assumptions the transformer appears to demonstrate in practice are an interesting area of future research.

%% file: main.bbl
\begin{thebibliography}{30}
\providecommand{\natexlab}[1]{#1}
\providecommand{\url}[1]{\texttt{#1}}
\expandafter\ifx\csname urlstyle\endcsname\relax
  \providecommand{\doi}[1]{doi: #1}\else
  \providecommand{\doi}{doi: \begingroup \urlstyle{rm}\Url}\fi

\bibitem[Akyürek et~al.(2023)Akyürek, Schuurmans, Andreas, Ma, and Zhou]{akyürek2023learning}
Ekin Akyürek, Dale Schuurmans, Jacob Andreas, Tengyu Ma, and Denny Zhou.
\newblock What learning algorithm is in-context learning? investigations with linear models, 2023.

\bibitem[Bai et~al.(2023)Bai, Chen, Wang, Xiong, and Mei]{bai2023transformers}
Yu~Bai, Fan Chen, Huan Wang, Caiming Xiong, and Song Mei.
\newblock Transformers as statisticians: Provable in-context learning with in-context algorithm selection, 2023.

\bibitem[Bhattamishra et~al.(2020)Bhattamishra, Ahuja, and Goyal]{bhattamishra2020ability}
Satwik Bhattamishra, Kabir Ahuja, and Navin Goyal.
\newblock On the ability and limitations of transformers to recognize formal languages, 2020.

\bibitem[Bietti et~al.(2023)Bietti, Cabannes, Bouchacourt, Jegou, and Bottou]{bietti2023birth}
Alberto Bietti, Vivien Cabannes, Diane Bouchacourt, Herve Jegou, and Leon Bottou.
\newblock Birth of a transformer: A memory viewpoint.
\newblock In \emph{Thirty-seventh Conference on Neural Information Processing Systems}, 2023.

\bibitem[Brown et~al.(2020)Brown, Mann, Ryder, Subbiah, Kaplan, Dhariwal, Neelakantan, Shyam, Sastry, Askell, et~al.]{brown2020language}
Tom Brown, Benjamin Mann, Nick Ryder, Melanie Subbiah, Jared~D Kaplan, Prafulla Dhariwal, Arvind Neelakantan, Pranav Shyam, Girish Sastry, Amanda Askell, et~al.
\newblock Language models are few-shot learners.
\newblock \emph{Advances in neural information processing systems}, 33:\penalty0 1877--1901, 2020.

\bibitem[Dosovitskiy et~al.(2021)Dosovitskiy, Beyer, Kolesnikov, Weissenborn, Zhai, Unterthiner, Dehghani, Minderer, Heigold, Gelly, Uszkoreit, and Houlsby]{dosovitskiy2021image}
Alexey Dosovitskiy, Lucas Beyer, Alexander Kolesnikov, Dirk Weissenborn, Xiaohua Zhai, Thomas Unterthiner, Mostafa Dehghani, Matthias Minderer, Georg Heigold, Sylvain Gelly, Jakob Uszkoreit, and Neil Houlsby.
\newblock An image is worth 16x16 words: Transformers for image recognition at scale, 2021.

\bibitem[Edelman et~al.(2024)Edelman, Edelman, Goel, Malach, and Tsilivis]{edelman2024evolution}
Benjamin~L. Edelman, Ezra Edelman, Surbhi Goel, Eran Malach, and Nikolaos Tsilivis.
\newblock The evolution of statistical induction heads: In-context learning markov chains, 2024.

\bibitem[Giannou et~al.(2023)Giannou, Rajput, Sohn, Lee, Lee, and Papailiopoulos]{giannou23looped}
Angeliki Giannou, Shashank Rajput, Jy-Yong Sohn, Kangwook Lee, Jason~D. Lee, and Dimitris Papailiopoulos.
\newblock Looped transformers as programmable computers.
\newblock In \emph{Proceedings of the 40th International Conference on Machine Learning}, pages 11398--11442, 23--29 Jul 2023.

\bibitem[Goldowsky-Dill et~al.(2023)Goldowsky-Dill, MacLeod, Sato, and Arora]{goldowsky2023localizing}
Nicholas Goldowsky-Dill, Chris MacLeod, Lucas Sato, and Aryaman Arora.
\newblock Localizing model behavior with path patching.
\newblock \emph{arXiv preprint arXiv:2304.05969}, 2023.

\bibitem[He et~al.(2021)He, Chen, Xie, Li, Dollár, and Girshick]{he2021masked}
Kaiming He, Xinlei Chen, Saining Xie, Yanghao Li, Piotr Dollár, and Ross Girshick.
\newblock Masked autoencoders are scalable vision learners, 2021.

\bibitem[Hoogland et~al.(2024)Hoogland, Wang, Farrugia-Roberts, Carroll, Wei, and Murfet]{hoogland2024developmental}
Jesse Hoogland, George Wang, Matthew Farrugia-Roberts, Liam Carroll, Susan Wei, and Daniel Murfet.
\newblock The developmental landscape of in-context learning.
\newblock \emph{arXiv preprint arXiv:2402.02364}, 2024.

\bibitem[Li et~al.(2023)Li, Wang, Liu, and Chen]{li2023theoretical}
Hongkang Li, Meng Wang, Sijia Liu, and Pin-Yu Chen.
\newblock A theoretical understanding of shallow vision transformers: Learning, generalization, and sample complexity.
\newblock In \emph{The Eleventh International Conference on Learning Representations}, 2023.

\bibitem[Lin and Lee(2024)]{lin2024dual}
Ziqian Lin and Kangwook Lee.
\newblock Dual operating modes of in-context learning, 2024.

\bibitem[Liu et~al.(2023)Liu, Ash, Goel, Krishnamurthy, and Zhang]{liu2023transformers}
Bingbin Liu, Jordan~T. Ash, Surbhi Goel, Akshay Krishnamurthy, and Cyril Zhang.
\newblock Transformers learn shortcuts to automata, 2023.

\bibitem[Makkuva et~al.(2024)Makkuva, Bondaschi, Girish, Nagle, Jaggi, Kim, and Gastpar]{makkuva2024attention}
Ashok~Vardhan Makkuva, Marco Bondaschi, Adway Girish, Alliot Nagle, Martin Jaggi, Hyeji Kim, and Michael Gastpar.
\newblock Attention with markov: A framework for principled analysis of transformers via markov chains.
\newblock \emph{arXiv preprint arXiv:2402.04161}, 2024.

\bibitem[Nichani et~al.(2024)Nichani, Damian, and Lee]{nichani2024transformers}
Eshaan Nichani, Alex Damian, and Jason~D. Lee.
\newblock How transformers learn causal structure with gradient descent, 2024.

\bibitem[Norris(1997)]{norris1998markov}
J.~R. Norris.
\newblock \emph{Markov Chains}.
\newblock Cambridge Series in Statistical and Probabilistic Mathematics. Cambridge University Press, 1997.

\bibitem[Olsson et~al.(2022)Olsson, Elhage, Nanda, Joseph, DasSarma, Henighan, Mann, Askell, Bai, Chen, Conerly, Drain, Ganguli, Hatfield-Dodds, Hernandez, Johnston, Jones, Kernion, Lovitt, Ndousse, Amodei, Brown, Clark, Kaplan, McCandlish, and Olah]{olsson2022incontext}
Catherine Olsson, Nelson Elhage, Neel Nanda, Nicholas Joseph, Nova DasSarma, Tom Henighan, Ben Mann, Amanda Askell, Yuntao Bai, Anna Chen, Tom Conerly, Dawn Drain, Deep Ganguli, Zac Hatfield-Dodds, Danny Hernandez, Scott Johnston, Andy Jones, Jackson Kernion, Liane Lovitt, Kamal Ndousse, Dario Amodei, Tom Brown, Jack Clark, Jared Kaplan, Sam McCandlish, and Chris Olah.
\newblock In-context learning and induction heads, 2022.

\bibitem[Oymak et~al.(2023)Oymak, Rawat, Soltanolkotabi, and Thrampoulidis]{oymak2023attn-prompt}
Samet Oymak, Ankit~Singh Rawat, Mahdi Soltanolkotabi, and Christos Thrampoulidis.
\newblock On the role of attention in prompt-tuning.
\newblock In \emph{Proceedings of the 40th International Conference on Machine Learning}, 2023.

\bibitem[Pagliardini(2023)]{pagliardini-llm}
Matteo Pagliardini.
\newblock {GPT-2} modular codebase implementation.
\newblock \emph{https://github.com/epfml/llm-baselines}, 2023.

\bibitem[Pérez et~al.(2021)Pérez, Barceló, and Marinkovic]{perez2021turing}
Jorge Pérez, Pablo Barceló, and Javier Marinkovic.
\newblock Attention is {T}uring-complete.
\newblock \emph{Journal of Machine Learning Research}, 22\penalty0 (75):\penalty0 1--35, 2021.

\bibitem[Rajaraman et~al.(2024)Rajaraman, Jiao, and Ramchandran]{rajaraman2024theory}
Nived Rajaraman, Jiantao Jiao, and Kannan Ramchandran.
\newblock Toward a theory of tokenization in llms, 2024.

\bibitem[Sanford et~al.(2023)Sanford, Hsu, and Telgarsky]{NEURIPS2023_73bf6924}
Clayton Sanford, Daniel~J Hsu, and Matus Telgarsky.
\newblock Representational strengths and limitations of transformers.
\newblock In A.~Oh, T.~Naumann, A.~Globerson, K.~Saenko, M.~Hardt, and S.~Levine, editors, \emph{Advances in Neural Information Processing Systems}, volume~36, pages 36677--36707. Curran Associates, Inc., 2023.
\newblock URL \url{https://proceedings.neurips.cc/paper_files/paper/2023/file/73bf692447f174984f30499ec9b20e04-Paper-Conference.pdf}.

\bibitem[Sanford et~al.(2024)Sanford, Hsu, and Telgarsky]{sanford2024transformers}
Clayton Sanford, Daniel Hsu, and Matus Telgarsky.
\newblock Transformers, parallel computation, and logarithmic depth, 2024.

\bibitem[Vaswani et~al.(2017)Vaswani, Shazeer, Parmar, Uszkoreit, Jones, Gomez, Kaiser, and Polosukhin]{vaswani2017attention}
Ashish Vaswani, Noam Shazeer, Niki Parmar, Jakob Uszkoreit, Llion Jones, Aidan~N Gomez, {\L}ukasz Kaiser, and Illia Polosukhin.
\newblock Attention is all you need.
\newblock In \emph{Advances in Neural Information Processing Systems}, pages 5998--6008, 2017.

\bibitem[Wei et~al.(2022{\natexlab{a}})Wei, Chen, and Ma]{wei2022turing-approx}
Colin Wei, Yining Chen, and Tengyu Ma.
\newblock Statistically meaningful approximation: a case study on approximating {T}uring machines with transformers.
\newblock In \emph{Advances in Neural Information Processing Systems}, volume~35, pages 12071--12083, 2022{\natexlab{a}}.

\bibitem[Wei et~al.(2022{\natexlab{b}})Wei, Wang, Schuurmans, Bosma, Xia, Chi, Le, Zhou, et~al.]{wei2022chain}
Jason Wei, Xuezhi Wang, Dale Schuurmans, Maarten Bosma, Fei Xia, Ed~Chi, Quoc~V Le, Denny Zhou, et~al.
\newblock Chain-of-thought prompting elicits reasoning in large language models.
\newblock \emph{Advances in neural information processing systems}, 35:\penalty0 24824--24837, 2022{\natexlab{b}}.

\bibitem[Weiss et~al.(2021)Weiss, Goldberg, and Yahav]{weiss2021thinking}
Gail Weiss, Yoav Goldberg, and Eran Yahav.
\newblock Thinking like transformers.
\newblock In \emph{International Conference on Machine Learning}, pages 11080--11090, 2021.

\bibitem[Yao(1979)]{yao1979some}
Andrew Chi-Chih Yao.
\newblock Some complexity questions related to distributive computing (preliminary report).
\newblock In \emph{Proceedings of the eleventh annual ACM symposium on Theory of computing}, pages 209--213, 1979.

\bibitem[Yun et~al.(2020)Yun, Bhojanapalli, Rawat, Reddi, and Kumar]{Yun2020seq2seq}
Chulhee Yun, Srinadh Bhojanapalli, Ankit~Singh Rawat, Sashank Reddi, and Sanjiv Kumar.
\newblock Are transformers universal approximators of sequence-to-sequence functions?
\newblock In \emph{International Conference on Learning Representations}, 2020.

\end{thebibliography}
